\documentclass[10pt,twocolumn,letterpaper]{article}

\usepackage[pagenumbers]{cvpr}

\usepackage[pagebackref,breaklinks,colorlinks,allcolors=cvprblue]{hyperref}
\usepackage{mathtools, amsmath, amssymb}
\usepackage{siunitx, kotex, tabularx, array, xcolor, pifont, makecell, multirow, booktabs}
\newcommand{\cmark}{\textcolor{green!60!black}{\ding{51}}}
\newcommand{\xmark}{\textcolor{red!70!black}{\ding{55}}}

\usepackage{multirow}

\newcommand{\subpara}[1]{%
  \vspace{0.5em}
  \noindent\textbf{#1}
}

\definecolor{cvprblue}{rgb}{0.21,0.49,0.74}
\def\confName{CVPR}
\def\confYear{2026}

\title{CulinaryCut-VLAP: A Vision–Language–Action–Physics Framework \\ for Food Cutting via a Force-Aware Material Point Method}
\author{
Hyunseo Koh$^{1}$ \quad
Chang-Yong Song$^{2}$ \quad
Youngjae Choi$^{1}$ \quad
Misa Viveriros$^{2}$ \\[0.6em]
David Hyde$^{2}$ \quad
Heewon Kim$^{1}$ \\[0.8em]
$^{1}$Soongsil University \quad $^{2}$Vanderbilt University \\[0.5em]
{\tt\small\{deepvelop, yj951118, hwkim\}@soongsil.ac.kr}\\
{\tt\small\{chang-yong.song, misa.h.viveiros, david.hyde.1\}@vanderbilt.edu}
}

\begin{document}
\maketitle

\begin{abstract}
    Food cutting is a highly practical yet underexplored application at the intersection of vision and robotic manipulation. The task remains challenging because interactions between the knife and deformable materials are highly nonlinear and often entail large deformations, frequent contact, and topological change, which in turn hinder stable and safe large-scale data collection.    
    To address these challenges, we propose a unified framework that couples a vision-language-action (VLA) dataset with a physically realistic cutting simulator built on the material point method (MPM). Our simulator adopts MLS-MPM as its computational core, reducing numerical dissipation and energy drift while preserving rotational and shear responses even under topology-changing cuts. During cutting, forces and stress distributions are estimated from impulse exchanges between particles and the grid, enabling stable tracking of transient contact forces and energy transfer.
    We also provide a benchmark dataset that integrates diverse cutting trajectories, multi-view visual observations, and fine-grained language instructions, together with force--torque and tool--pose labels
    to provide physically consistent training signals.
    These components realize a learning--evaluation loop that respects the core physics of cutting and establishes a safe, reproducible, and scalable foundation for advancing VLA models in deformable object manipulation.
\end{abstract}
\begin{table*}[t!]
\scriptsize
\centering
\begin{tabular}{l|c c c c c c c c c c}
\hline
\textbf{Dataset} &
\makecell{\textbf{Physics}\\\textbf{Sim}} &
\makecell{\textbf{Robot}\\\textbf{Sim}} &
\makecell{\textbf{Data}\\\textbf{Origin}} &
\makecell{\textbf{Multi}\\\textbf{Camera}} &
\makecell{\textbf{Language}\\\textbf{Intruction}} &
\makecell{\textbf{Continuous}\\\textbf{Action}} &
\makecell{\textbf{Cutting}\\\textbf{Flexibility}} &
\makecell{\textbf{Cutting}\\\textbf{Task}} &
\makecell{\textbf{Multiple cut}\\\textbf{Style}} &
\makecell{\textbf{Force}\\\textbf{Data}}\\
\hline
DROID~\cite{droid2024} & N/A & N/A & Real & \cmark & \cmark & \cmark & \xmark & \xmark  & \xmark & \xmark\\
FMB~\cite{luo2025fmb} & N/A & N/A & Real & \cmark & \xmark & \xmark & \xmark & \xmark & \xmark & \xmark\\
Playing with Food~\cite{sawhney2020playing} & N/A & N/A & Real & \cmark & \xmark & \xmark & \xmark & \xmark & \xmark & \xmark\\
MimicPlay~\cite{wang2023mimicplay} & N/A & MuJoCo & Real & \cmark & \xmark & \cmark & \xmark & \xmark & \xmark & \xmark\\
Mutex~\cite{shah2023mutex} & N/A & N/A & Real & \cmark & \xmark & \cmark & \xmark & \xmark & \xmark & \xmark\\
FurnitureBench~\cite{heo2025furniturebench} & N/A & IsaacGym~\cite{makoviychuk2021isaacgymhighperformance} & Real & \cmark & \xmark & \xmark & \xmark & \xmark & \xmark & \xmark\\
Arnold~\cite{gong2023arnold} & N/A & IssacSim & Synthetic & \cmark & \xmark & \cmark & \xmark & \xmark & \xmark & \xmark\\ 
Robomind~\cite{wu2025robomindbenchmarkmultiembodimentintelligence} & N/A & IssacSim & Synthetic & \cmark & \xmark & \cmark & \xmark & \xmark  & \xmark & \xmark\\ 
\hline
RoboChop~\cite{dikshit2023robochop} & N/A & N/A & Real & \xmark & \xmark & \xmark & \cmark & \cmark & \xmark & \xmark\\
RoboCook~\cite{shi2023robocook} & MPM & N/A & Hybrid & \cmark & \xmark & \xmark & \cmark & \cmark & \cmark & \xmark\\
Foo et al.~\cite{jamdagni2021robotic} & FEM & N/A & Synthetic & \xmark & \xmark & \xmark & \cmark & \cmark & \xmark & \xmark\\
SliceIt~\cite{sliceit2024} & FEM & \text{Gazebo} & Synthetic & \xmark & \xmark & \xmark & \cmark & \cmark & \xmark & \xmark\\
DiSECt~\cite{disect2022} & FEM & N/A & Synthetic & \cmark & \xmark & \xmark & \cmark & \cmark & \xmark & \xmark\\
TopoCut~\cite{topocut2025} & MPM & N/A & Synthetic & \cmark & \xmark & \xmark & \cmark & \cmark & \cmark & \xmark\\
Ours & MPM & \text{Maniskill}~\cite{tao2025maniskill3gpuparallelizedrobotics} & Synthetic & \cmark & \cmark & \cmark & \cmark & \cmark & \cmark & \cmark\\
\hline
\end{tabular}
\caption{\textbf{Comparison of cutting benchmark datasets for embodied agents.}~\textbf{Physics Sim:} Realistic physical simulation using a physics-based simulator. \textbf{Robot Sim:} Ability to perform actions interactively within a robot simulator. \textbf{Data Origin:} Source of collected data (real-world, synthetic, or hybrid). “Hybrid” indicates a combination of real and synthetic data. \textbf{Multi Camera:} Robot is equipped with multiple cameras for perception. \textbf{Language:} Task goals are specified using human language instructions. \textbf{Continuous Action:} Actions are defined in a continuous control space rather than discrete steps. \textbf{Cutting Flexibility:} Support for diverse geometric and topological deformations from cutting.
\textbf{Cutting Task:} Inclusion of cutting-related tasks. \textbf{Multiple Cut Style:} Support for multiple cutting styles or multi-step cutting sequences. \textbf{Force Data:} Availability of force-sensing data.}
\label{relatework:Tab}
\end{table*}
\section{Introduction}
\label{sec:intro}
\begin{figure}
    \centering
    \includegraphics[width=0.7\linewidth]{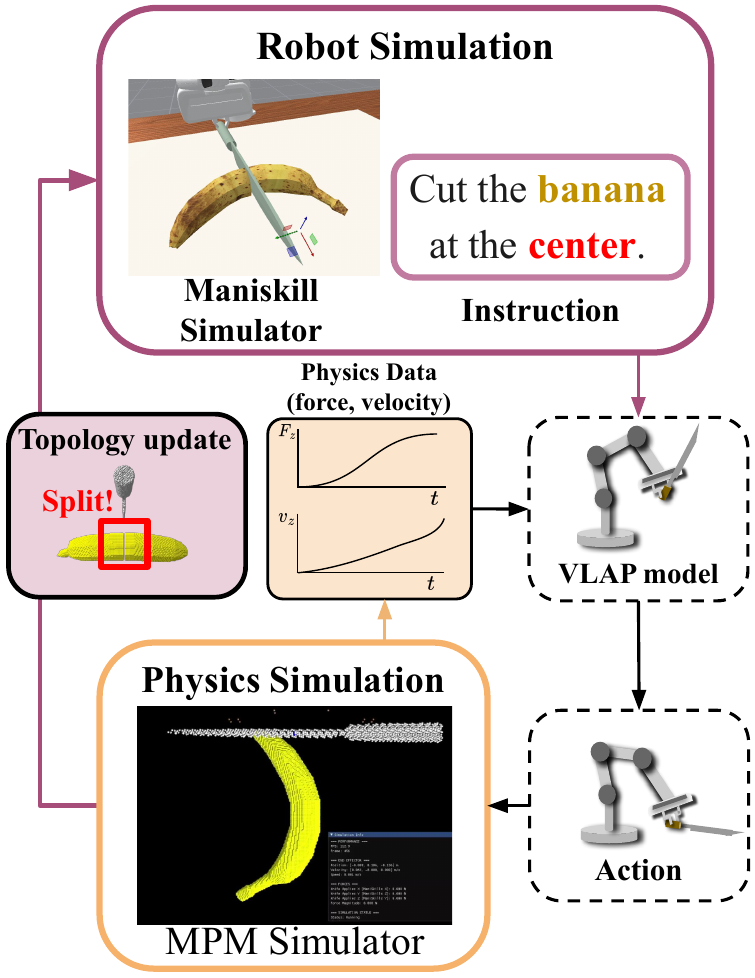}
    \caption{Concept of the Vision–Language–Action-Physics (VLAP) framework. The framework interacts between the robot and physics simulation to construct a physically grounded VLA dataset and model for food cutting.}
    \label{fig:placeholder}
    \vspace{-1em}
\end{figure}
Interest in vision-based robotic manipulation~\cite{openvla,liu2025rdt1bdiffusionfoundationmodel} has grown rapidly with advances in low-cost vision sensors, sim-to-real learning~\cite{yang2023learning}, and the integration of perception~\cite{xia2018gibson}, language~\cite{achiam2023gpt4}, and control~\cite{niu2024screenagentvisionlanguagemodeldriven}.
Recent VLA models~\cite{bu2025univla,dreamvla25,openvla} unify these components into a single language→perception→action pipeline trained under physical constraints~\cite{yu2025forcevla}, enabling operation in grounded environments.

Despite this progress, food cutting remains largely unexplored in VLA research. Cutting tasks involve deformation, fracture, and force-mediated contact—phenomena~\cite{disect2022} that rigid-body manipulation datasets fail to capture.
Existing approaches rely either on real robot data, which ensures physical realism but limits scale, or on geometry-focused simulations~\cite{topocut2025} that do not directly leverage forces and impulses, which can potentially limit physical accuracy.

Constructing a dataset suitable for VLA-based food cutting introduces three challenges:~(1)~existing robot simulators used in VLA frameworks are limited in modeling topology changes and force/velocity variations during deformable interactions.
(2)~cutting outcomes are continuous in size, ratio, and orientation, requiring large-scale diverse data, and
(3)~multi-step interactions tend to generate a wide semantic gap between language instructions and quantitative outcomes;
and (4)~ensuring safety-aware control in cutting, where force constraints must be respected to prevent unstable or unsafe trajectories.

Concurrently, advances in robot hardware and high-fidelity simulation (e.g., MPM/FEM) enable large-scale, physically consistent deformable-manipulation data~\cite{Xu_2025} for policy learning and sim-to-real transfer.
In particular, MLS-MPM\citep{2018-MLSMPM} handles contact-rich, topology-changing phenomena while reducing numerical diffusion and energy drift and preserving rotational and shear responses—properties that are crucial for cutting.

Nevertheless, large-scale multimodal datasets suitable for VLA training remain scarce, especially for cutting\cite{openxembodiment2023}, where safety and data-collection constraints make real-world scaling impractical~\cite{khazatsky2024droidlargescaleinthewildrobot}. These challenges are compounded by the nature of cutting, which demands precise goal specification via quantitative spatial instructions (exact positions, orientations, division ratios). Commands such as “cut through the apple’s center,” “slice the cucumber into three equal parts,” require tight coupling of perception, spatial reasoning, and force control, posing a quantitative grounding challenge: mapping language, with numeric precision, into executable trajectories.

To bridge these gaps, we present a unified benchmark that couples a VLA dataset for precise grounding of quantitative instructions with an MPM-based cutting simulator. The simulator uses MPM to stably reproduce contact-rich, topology-changing cutting interactions and to estimate force/stress distributions from impulse exchanges between particles. It includes a fracture criterion with particle/mesh updates, frictional knife–material contact via CPIC for stable momentum exchange, and safety controllers (force/velocity caps, entry-angle guards) to regularize contact transients. Sim-to-real consistency is validated via force–time curves and cutting-surface quality metrics.

The dataset encodes both qualitative intent and quantitative cutting goals (position, orientation, division ratio), enabling large-scale multimodal learning and sim-to-real transfer for deformable-object cutting within a physically consistent framework.

In summary, the present work contributes:
\begin{itemize}
\item A \textbf{large-scale food cutting dataset} providing diverse food categories and cutting styles with multimodal annotations for quantitative instruction grounding.
\item A \textbf{hybrid simulation framework} combining ManiSkill and MPM to model deformation, contact forces, and topology changes under realistic visual and physical conditions.
\item A \textbf{scalable data generation pipeline} using LLM-based instruction synthesis and simulation-driven augmentation for efficient large-scale dataset construction.
\item A \textbf{comprehensive benchmark and analysis} of multiple VLA models, revealing challenges in quantitative grounding, continuous action control, and generalization.
\end{itemize}
\begin{figure*}[t!]
    \centering
    \includegraphics[width=\linewidth]{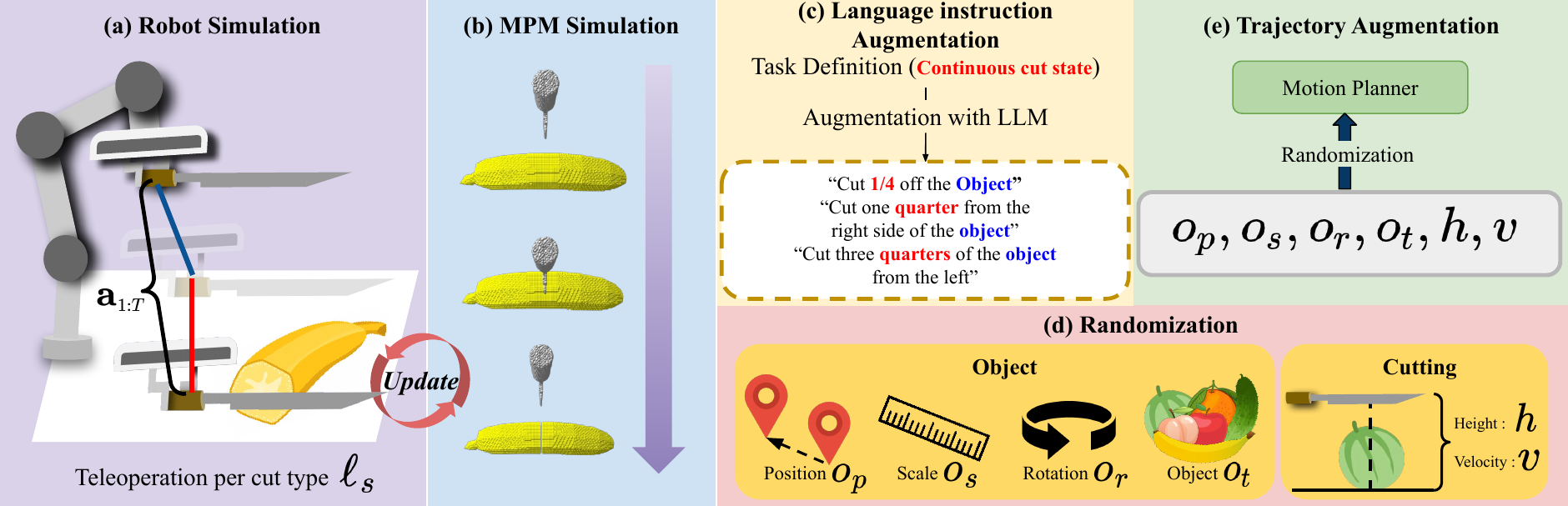}
    \caption{Overview of CulinaryCut data generation pipeline.
(a) Teleoperation generates initial cutting demonstrations for each cut style.
(b) Physics simulation updates material deformation and cutting geometry.
(c) Language instructions are augmented using LLMs with continuous cut state variations.
(d, e) Trajectories are augmented via motion planning with object and cutting randomization.}
\vspace{-1em}
\end{figure*}
\section{Related Work}
\label{sec:related work}
\paragraph{Large multimodal models (LMMs).}
Recent advances in LMMs extend text-only LLM reasoning to visual and temporal domains by aligning frozen vision encoders with open-weight LLMs through lightweight adapters, enabling multimodal instruction-following and dialogue~\cite{achiam2023gpt4,radford2021clip,liu2023llava,zhu2023minigpt4}.
Region- and video-aware variants enhance spatial and temporal grounding via RoI features and interleaved visual--text tokens~\cite{cai2023vipllava,rasheed2024glamm,li2024llavainterleave,fang2024mmbenchvideo}, while benchmarks such as MMBench and SEED-Bench systematically evaluate these multimodal capabilities~\cite{liu2024mmbench,li2023seedbench2}.
Building on such perception-aligned foundations, reasoning-oriented frameworks incorporate interactive correction, memory, and deliberative intermediates to support long-horizon task understanding and multi-step decision making~\cite{huang2023instruct2actmappingmultimodalityinstructions,lynch2022interactivelanguagetalkingrobots,wu2025foresightforethoughtvlminthelooppolicy,zhang2025inspire}.
Recent systems elevate LLMs into full perception--planning--control orchestrators~\cite{geminiroboticsteam2025geminiroboticsbringingai,yang2025agenticrobotbraininspiredframework}, while multimodal reasoning extends beyond language to speech and haptics for fine-grained embodied control~\cite{huang2025tactilevlaunlockingvisionlanguageactionmodels,khan2025shakevlavisionlanguageactionmodelbasedbimanual}, pointing toward unified, adaptive, and deployable generalist agents.

\paragraph{Vision-language-action (VLA) for robotics.}
VLA models unify perception, language, and action generation into a single policy backbone, progressing from early generalist tokenization to large-scale real-world transformers and open cross-embodiment architectures~\cite{reed2022generalistagent,brohan2023rt2visionlanguageactionmodelstransfer,kim2024openvlaopensourcevisionlanguageactionmodel}.
Reasoning-augmented VLAs build upon LMMs by integrating self-talk, VLM$\rightarrow$VLA transfer, and platform-level orchestration for long-horizon tasks~\cite{huang2022innermonologueembodiedreasoning,li2024visionlanguage,geminiroboticsteam2025geminiroboticsbringingai}.
Trajectory-centric designs map language to motion/control tokens and leverage video-pretrained world models for predictive control~\cite{bucker2022lattelanguagetrajectorytransformer,jiang2023vimageneralrobotmanipulation,wu2024unleashing}.
Applications span bimanual/mobile manipulation and driving/aerial domains with 3D sensing enriching spatial grounding~\cite{gbagbe2024bivlavisionlanguageactionmodelbasedbimanual,zhou2025opendrivevlaendtoendautonomousdriving,li2025pointvlainjecting3dworld}.
Efficiency and scalability are achieved through hierarchical decomposition, token-efficient actionization, and heterogeneous-robot pretraining with model--expert collaboration~\cite{ajay2023compositionalfoundationmodelshierarchical,pertsch2025fastefficientactiontokenization,xiang2025vlamodelexpertcollaborationbidirectional,zhang2025upvlaunifiedunderstandingprediction}.

\paragraph{Large-scale robot learning datasets.}
Community efforts have established broad open datasets for imitation and VLA-style policies, 
yet few directly address deformable cutting. 
Open X-Embodiment (RT-X) unifies over one million real-robot trajectories from 60 datasets across 34 labs and 22 embodiments~\citep{openxembodiment2023}, 
while RoboNet~\citep{robonet2019}, BridgeData~V2~\citep{bridgedata2023}, DROID~\citep{droid2024}, and LIBERO~\citep{libero2023} provide large-scale manipulation datasets but exclude deformable interactions. 
Cutting-specific resources remain simulation-only~\citep{disect2022,sliceit2024,topocut2025}, and force–deformation datasets~\citep{yu2025forcevla} are still lacking, as summarized in Table~\ref{relatework:Tab}.
Our dataset bridges this gap by introducing Set-of-Mark (SoM) and Trace-of-Mark (ToM) supervision, 
explicitly encoding where to cut and how the tool evolves over time to enable fine-grained spatial--temporal reasoning for deformable cutting.

\paragraph{Material point method (MPM).}
The material point method (MPM)~\citep{sulsky1994particle} excels at simulating solid mechanics involving large deformations and fractures~\citep{2013-mpm, 2019-cdmpm, 10.1145/3340259}. As a hybrid method coupling Lagrangian particles with an Eulerian grid, MPM effectively handles material deformation and topological change. Standard PIC/FLIP transfer methods used in early MPM, however suffered from severe numerical dissipation, failing to conserve rotational momentum. To address this dissipation problem, the Affine Particle-in-Cell (APIC) technique~\citep{jiang2017angular} was proposed, which effectively preserves rotational and shear stress via the affine velocity term $\mathbf{C}$.
Our work adopts MLS-MPM (moving least-squares MPM)~\citep{2018-MLSMPM}, which incorporates the benefits of APIC, as our computational core to ensure a stable simulation foundation for synthetic data generation.
\section{CulinaryCut Benchmark}
\label{sec:Benchmark}
To evaluate the robot’s food cutting performance under text instructions, we introduce a cutting task benchmark in a hybrid simulation for the robot and physics. 

\subsection{Robot Simulation Environment}
\paragraph{Simulation Platform.}  
We adopt ManiSkill~\cite{tao2025maniskill3gpuparallelizedrobotics}, for robot simulation, which provides accurate robot movement and a visually realistic environment with three viewpoints of front, left, and right.
It detects the moment when the knife first contacts the object surface and marks it as the contact phase.
We integrate the topology change from the physics simulation during the contact phase into ManiSkill.

\paragraph{Robotic Manipulation.}  
For task execution, we utilize a 7-DoF Franka Emika Panda robot, equipped with a cutting knife as the end effector. 
The agent directly controls all joints and the gripper using ManiSkill’s built-in controller for stable and precise manipulation. 
End-effector actions are represented in 3D translation and quaternion rotation, with optional proportional derivative (PD) joint representations for different types of action model training. 

\subsection{Physics Simulation Environment}
\label{Subsec:Physics Simulation}

\paragraph{Simulation Platform.}
We employ the moving least-squares material point method (MLS-MPM)~\cite{2018-MLSMPM}, which is known to be effective for cutting simulation~\cite{2018-MLSMPM}. The knife and cutting board are modeled as signed distance fields (SDFs), enabling efficient contact detection with MPM-based deformable objects.
Full simulation method details are in the supplementary material (Section~\ref{sec:mpm}).

\paragraph{Topology Change.} Each particle is associated with a continuum damage scalar \(D \in [0,1]\). At each time step, particles near the knife blade receive increased values of \(D\), which linearly reduce their effective moduli of Lam\'e and promote crack formation. 

\paragraph{Force and Velocity Calculation.}
After resolving contact interactions on the grid, we compute the impulse of the knife during the time step by comparing the velocities before and after contact, from which the applied cutting force is derived. 
We additionally track the blade velocity and adopt a quadratic speed-resistance model to ensure that cutting occurs only when the blade moves sufficiently fast.

\subsection{Task Definition}
\label{Subsec:task_define}
\paragraph{Cut Styles.}
We define four primary \textbf{cut styles}~($\ell_s$) that characterize how the robot interacts with the object during cutting: 
\textit{Normal Cut}, \textit{Bias Cut}, \textit{Guillotine Cut}, and \textit{Saw Cut}.  
Normal Cut represents a straight cut with a perpendicular slicing motion.  
Bias Cut introduces an angled cutting trajectory, commonly used to increase contact area or create oblique slices.  
Guillotine Cut represents a downward slicing motion with the knife tip fixed, moving straight down along a single, controlled path.
Saw Cut involves repeated back-and-forth motions along the cutting direction, capturing more realistic interactions with more fibrous materials.

\paragraph{Continuous Cut States.}
We describe the progress and goal of each cutting operation using \textbf{continuous cut states}: \textit{Middle Cut}, \textit{Split Cut}, and \textit{Ratio Cut}.  
Middle Cut denotes cuts targeting the geometric center or midpoint of the object.  
Split Cut corresponds to \textbf{multi-cut} operations that divide the object into multiple segments (\textit{e.g.}, slicing a cucumber into several pieces).  
Ratio Cut specifies cuts at a desired proportion of the object, such as ``cut at one-third from the left'' or ``slice at 70\% of the length.''

\paragraph{Cutting Tasks.}
Each cutting task in our dataset is defined by a combination of a \textit{cut style} and a \textit{continuous cut state}.  
In addition, the language instruction explicitly specifies both the target state (\textit{e.g.}, numerical ratio or number of segments) and the target object, ensuring that the robot receives precise, goal-directed commands.  
For example, a task may require a Saw Cut with a Split Cut state (multi-step slicing into several pieces), or a Normal Cut with a Ratio Cut state (single slice at a specified ratio).  
This formulation enables the VLA model to jointly reason about \textit{how} to cut (style) and \textit{where/how much} to cut (state), allowing for fine-grained control over realistic food cutting behaviors.

\paragraph{Success Criteria.}  
A trial is considered successful if the resulting cut position lies within a positional tolerance of one-tenth of the object’s total length.  
In addition to positional accuracy, the cutting motion must preserve a correct blade orientation, ensuring that the contact angle between the knife and the target plane remains within the same tolerance range.  
This criterion guarantees that the robot not only cuts at the correct location but also performs a spatially aligned and directionally consistent motion, while allowing minor variations from simulation dynamics.

\subsection{Data Collection}
\label{Subsec:Data Collection}

\paragraph{Human Demonstration.}
For each cut style, a human operator provides teleoperated demonstrations within the simulation environment.  
During teleoperation, the relative pose of the knife with respect to the target object is recorded and later used for trajectory augmentation.

\paragraph{Trajectory Augmentation.}
A motion planner adapts each human-provided demonstration to various object conditions, including position~($O_p$), scale~($O_s$), rotation~($O_r$), object type~($O_t$), and cutting parameters of height~($\mathcal{h}$) and velocity~($\mathcal{v}$).  
These conditions and parameters are randomly sampled for data augmentation.
The planner generates realistic trajectories by computing the object's Axis-Aligned Bounding Box (AABB) to estimate its spatial extent and determining the appropriate cutting position based on the given cut style~($\ell_s$) and state. 

\paragraph{Topology and Physics Data.}
During trajectory execution, the physics simulation engine continuously updates the tool--object interaction by applying realistic physical forces and handling state transitions.  
The engine also updates the object's topology to reflect deformation and separation as cutting progresses.  
By incorporating these physics-driven topology changes, the system maintains coherent alignment between visual observations and manipulation actions, enabling physically grounded supervision.

\paragraph{Language Instructions.}  
For each trajectory, we utilized a template-based language generator to create diverse text instructions that described the same manipulation. 
Each cut state included 5+ templates with placeholders for object and continuous state terms, filled with varied lexical candidates (\textit{e.g.}, ``cut the banana at 0.5 ratio from the right side'', ``slice the banana halfway along its length''). 
A pool of equivalent expressions (\textit{e.g.}, ``50\%'', ``half'', ``two quarters'') was randomly substituted to enhance linguistic diversity. 
Since the instruction omits the initial state, the agent must infer the current scene from visual input to execute the command correctly. 
All templates and lexical variations are listed in the supplementary material.

\paragraph{Data Statistics.}
The CulinaryCut dataset consists of 325,000 simulation-based manipulation trajectories, where each task is defined as a unique combination of a \textit{cut style}~($\ell_s$) and a \textit{cut state}. We include five representative cut styles and thirteen cut states, composed of nine ratio cuts ranging from 0.1 to 0.9, one middle cut, and three split cuts (3-way, 4-way, and 5-way splits). For every cutting task, the system generates 500 augmented trajectories using a motion planner, and the remaining elements of the scene are randomly sampled.
The dataset comprises seven food categories~($\O_t$): orange, strawberry, melon, cucumber, banana, apple, and peach. All trajectories are collected across two distinct scenes, a realistic kitchen environment and a standard table setting, reflecting common everyday cutting contexts. Each trajectory is paired with more than five language instructions generated and augmented by large language models that cover a broad spectrum of numerical expressions, ratio descriptions, spatial directives, and phrasing styles. We allocate 20 trials per cutting task for testing, and the remaining are used for training.

\paragraph{Evaluation Setting.}
We design three evaluation environment settings to comprehensively assess model generalization:  
(1) evaluation under unseen random seeds involving novel object positions and scales not used during training.
(2) multi-object evaluation where multiple objects coexist within the same workspace, and  
(3) evaluation using unseen language instructions specifically reserved for testing.  
Additionally, user-defined cross-domain evaluation is supported, 
where models only trained on a specific object category (e.g., apple) can be evaluated on unseen categories (e.g., banana) to measure compositional generalization.

\section{CulinaryCut-VLAP}
\begin{figure}
    \centering
    \includegraphics[width=0.7\linewidth]{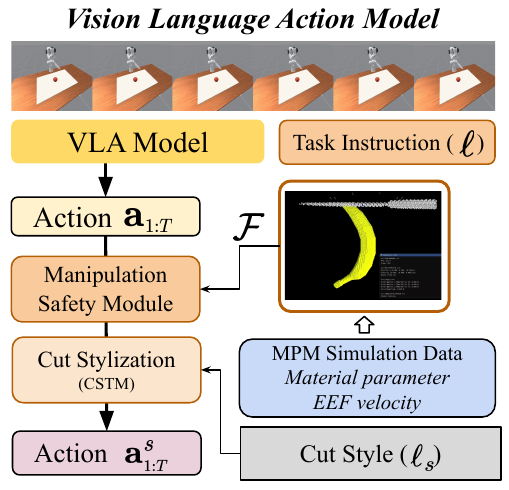}
    \caption{Overall pipeline of the \textbf{Vision-Language-Action-Physics (VLAP)} model during inference on the \textbf{CulinaryCut} dataset.}
    \label{fig:placeholder}
    \vspace{-1em}
\end{figure}
\subsection{Vision-Language-Action Model}
Our baseline follows the sequence-level Vision-Language-Action~(VLA) formulation adopted in large-scale models such as OpenVLA~\cite{openvla}, Octo~\cite{octomodelteam2024octoopensourcegeneralistrobot}, and RDT-1B~\cite{liu2025rdt1bdiffusionfoundationmodel}. 
Given a language instruction $\ell$ and a multimodal observation sequence $o_{1:T} = \{(I_t, x_t)\}_{t=1}^T$, 
where $I_t$ denotes multi-view RGB images and $x_t$ represents proprioceptive states, the VLA model $\pi_\theta$ generates an entire action sequence, 
\begin{equation}
    \hat{\mathbf{a}}_{1:T} = \pi_\theta(o_{1:T}, \ell) .
\end{equation}
%
\subsection{Manipulation Safety Module}
\label{Subsec:Safety Module}
To prevent excessive forces beyond the robot’s mechanical limits, we introduce a manipulation safety module that regulates knife velocity based on force predictions from the physics simulator. 
In the Franka robot, the contact force parameter ($F_{\max}$) above $100\,\mathrm{N}$ may cause hardware damage, making velocity control critical for safe manipulation.
From the simulation, we construct a dataset 
$\mathcal{D} = \{ (v_i, \mathbf{m}_i, F_i) \}_{i=1}^N$, 
where $v_i$ is the velocity of the knife, $\mathbf{m}_i$ denotes the properties of the material (\textit{e.g.}, $E$, $\sigma_y$), and $F_i$ is the resulting contact force. 
A regression model $\mathcal{R}$ predicts expected maximum force:
\begin{equation}
\hat{F} = \mathcal{R}(v, \mathbf{m}),
\end{equation}
and the safe velocity threshold is defined by
\begin{equation}
v_{\mathrm{safe}} = 
\max_{v} \;\text{s.t.}\; \mathcal{R}(v, \mathbf{m}) \le F_{\max},
\end{equation}
ensuring that the forces remain within the safety limits. 
This data-driven limiter integrates with the compliance controller, allowing a material-aware adaptive cutting motion.
\begin{figure}[t!]
    \centering
    \includegraphics[width=\linewidth]{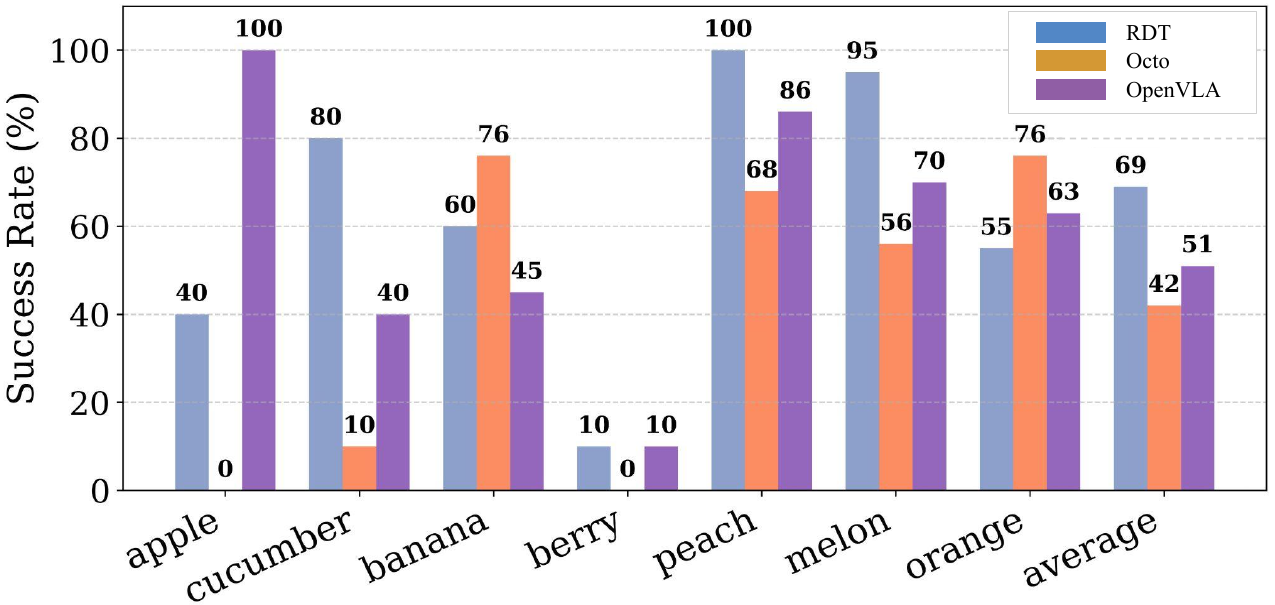}
    \caption{\textbf{Object Variation Results.}~The bar chart shows the inference performance of each model when trained on individual object–instruction pairs.}
    \label{fig:object_var}
\end{figure}
\subsection{Cutting Style Transfer Module (CSTM)}
CSTM transforms the action generated by VLA into the action generated by auxiliary networks. 

A ViT-based binary classifier~$\mathcal{C}$ predicts the contact state between the knife and objects from observation images $I_t$. 

Then, a cut-style generator~$f$ converts the predicted trajectory~$\hat{\mathbf{a}}_{1:T}$ into a style-specific trajectory.
Built upon behavior cloning, the module uses the contact segments of normal cutting trajectories~$\mathbf{a}$ and their style-specific counterparts~$\mathbf{a}^s$ as supervision, 
enabling VLA outputs to be adapted to different styles. 
Finally, the style-conditioned policy is defined as:
\[
\mathbf{a}^{\text{s}}_{1:T} =
\begin{cases}
f(\hat{\mathbf{a}}_{t:T}, \ell_{\text{s}}), & \text{if } \mathcal{C}(I_t) = 1, \\
\hat{\mathbf{a}}_{1:t}, & \text{otherwise},
\end{cases}
\]
where the generated stylized trajectory~$\mathbf{a}^{\text{s}}_{1:T}$ reflects both the linguistic intent~$\ell$ and the cut style instruction~$\ell_s$. 
\section{Experiments}
\label{sec:experiment}
\paragraph{Experimental Setup.}
All cutting experiments are conducted in a simulated tabletop environment with randomized object placements 
to evaluate each model's robustness and spatial generalization. 
Before each episode, the target object (e.g., banana, cucumber, apple, melon) is randomly initialized 
within the workspace boundary on the table. 
The robot executes the instructed cutting motion according to either a language-based or numerical command.
Each task is evaluated over 20 randomized trials (excluding the training seed), varying the object's initial pose and position to determine success.
\subsection{Experiment Results}
\begin{figure}[t!]
    \centering
    \includegraphics[width=\linewidth]{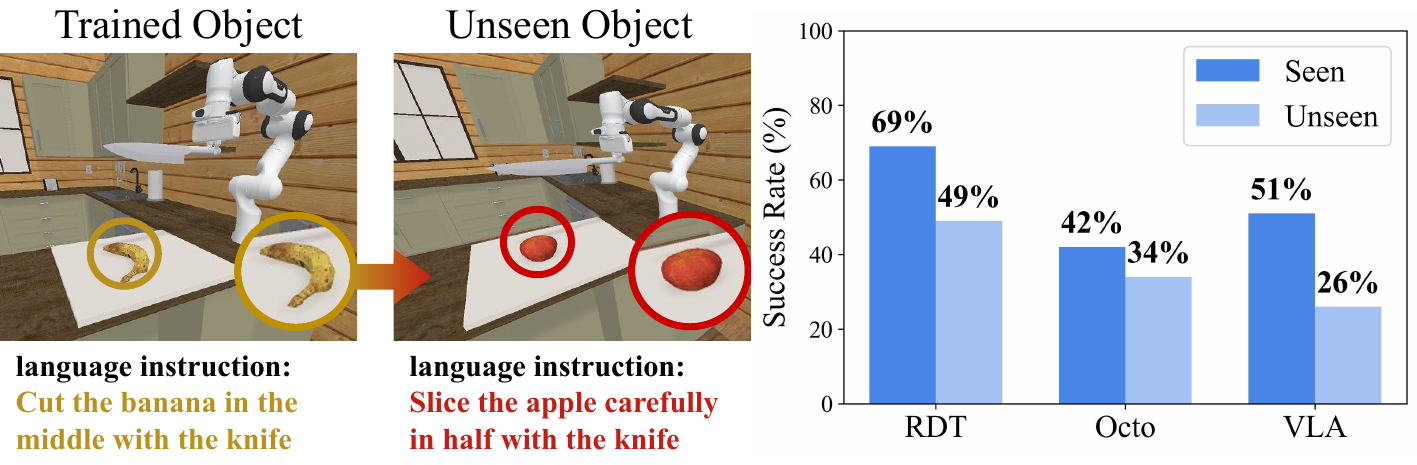}
    \caption{\textbf{Generalization to Unseen Objects.}~Evaluation of model generalization to unseen objects and novel language instructions. The bar chart presents success rates of RDT, Octo, and OpenVLA, showing a clear performance drop on unseen objects.}
    \label{fig:generalization_transfer}
    \vspace{-0.5em}
\end{figure}

\paragraph{General Task.}
As shown in Figure~\ref{fig:object_var}, models achieve stable performance when cutting normal-sized objects 
(e.g., \textit{banana}, \textit{cucumber}) at the half-cut ratio, 
but success rates drop sharply for smaller objects such as \textit{berry}, 
revealing limited precision and contact generalization at small scales.
\begin{figure}[t!]
    \centering
    \includegraphics[width=\linewidth]{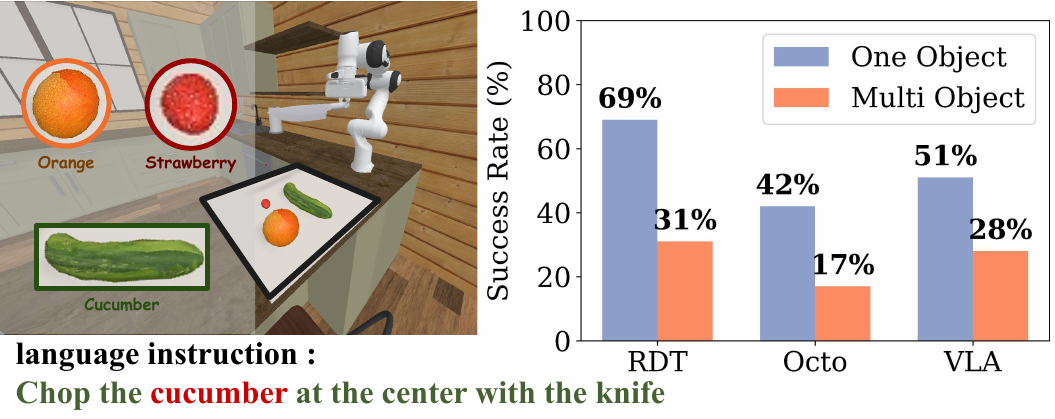}
    \caption{\textbf{Multi-Object Results:}~Comparison of RDT, Octo, and OpenVLA in single- and multi-object scenes. The presence of additional objects significantly impairs target identification and reduces cutting success rates.}
    \label{fig:multiobject_scene}
\end{figure}
\vspace{-6mm}
\paragraph{Multi-object Target Identification.}
To evaluate whether the model can correctly identify and cut the instructed object when multiple objects are present,
we conduct a multi-object scene experiment as illustrated in Figure~\ref{fig:multiobject_scene}.
Each scene contains multiple fruits placed randomly on the table,
and the model receives an instruction referring to one specific target (e.g., ``cut the banana at the center'').
The results, summarized in Figure~\ref{fig:multiobject_scene}, show a consistent performance drop across all models
when transitioning from single-object (Original) to multi-object scenarios.
Specifically, the success rate of RDT decreases from 68.57\% to 31.42\%,
Octo from 42.28\% to 17.00\%,
and OpenVLA from 50.57\% to 27.85\%.
This indicates that while the models possess moderate visual grounding ability,
their reasoning and spatial grounding for precise target selection
remain limited under complex, cluttered multi-object conditions.
\begin{figure*}[t!]
    \centering
    \includegraphics[width=\linewidth]{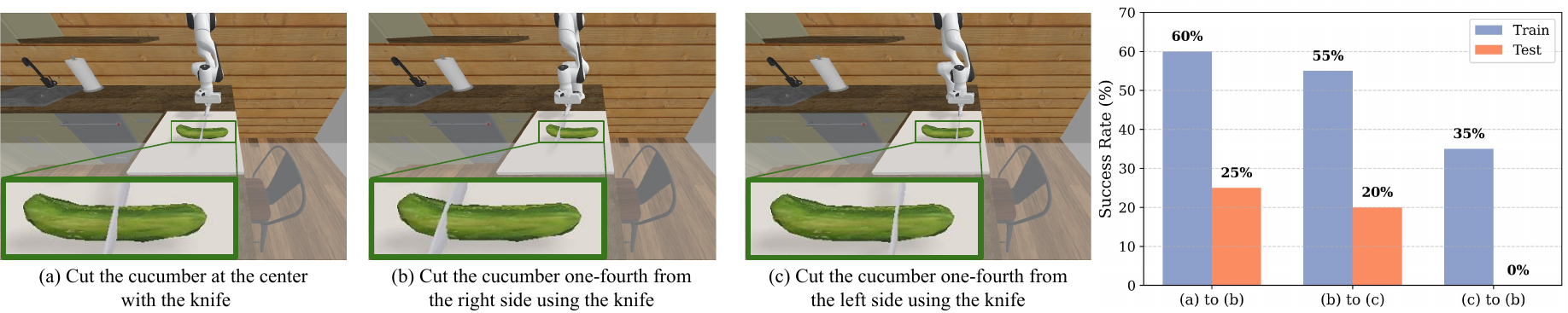}
    \caption{\textbf{Transfer Performance Results.}~(a), (b), and (c) illustrate ratio-based cutting tasks, while the bar chart on the right compares model performance between training and testing conditions, highlighting a clear performance drop on unseen ratio configurations.} 
    \label{fig:transfer}
\end{figure*}
\paragraph{Cross-object Generalization.}
We evaluate the baseline models' ability to generalize to unseen objects 
when trained on a single-object dataset and tested on different, untrained categories.
As shown in Figure~\ref{fig:generalization_transfer},
the performance exhibits a moderate drop when transferring to new objects:
RDT-1B decreases from 59.28\% to 48.14\%, and OpenVLA from 48.14\% to 45.00\%.
Although the success rates decline, both models still maintain reasonable performance 
on unseen objects, indicating that our dataset effectively promotes transferable 
cutting behavior beyond the trained object category.
This result suggests that the proposed dataset provides sufficient geometric and visual diversity 
to support generalizable cutting policies across object types.
\begin{figure}
    \centering
    \includegraphics[width=\linewidth]{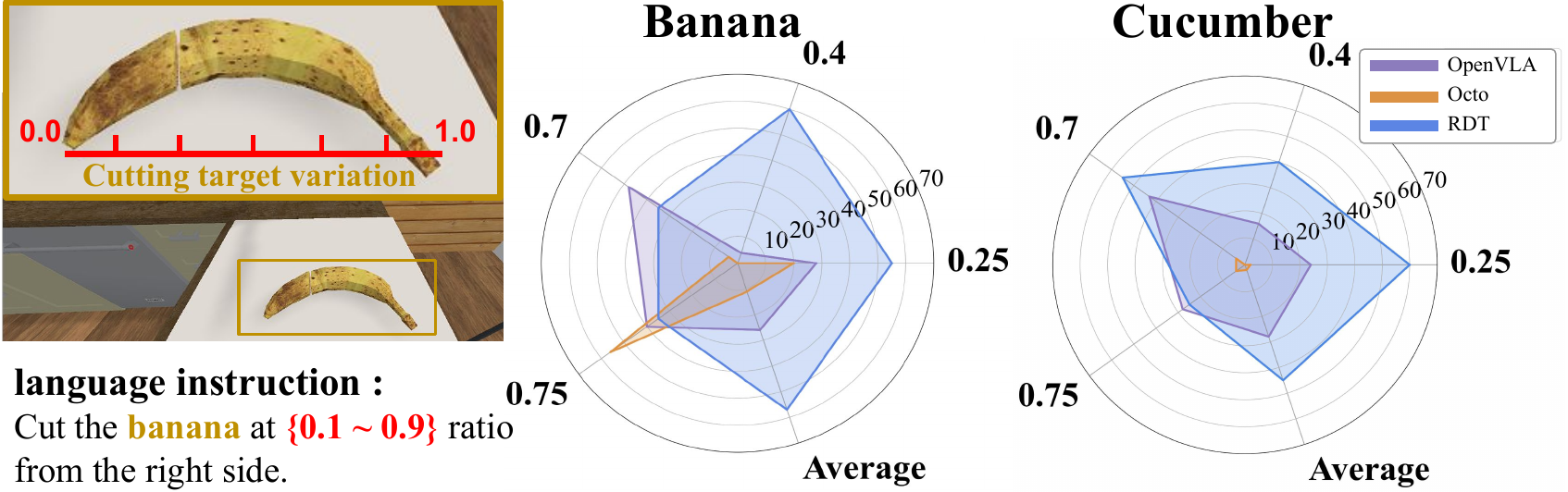}
    \caption{\textbf{Evaluation of Continuous Cut.}~Success rates (\%) of RDT, Octo, and OpenVLA under continuous ratio-based cutting instructions. The radar charts show model performance across cut ratios~(0.1–0.9) for banana and cucumber.}
    \label{fig:continuouscut}
    \vspace{-0.5em}
\end{figure}
%
\paragraph{Directional and Ratio-based Generalization.}
We further evaluate the RDT model, which achieved the highest overall performance, 
to analyze its ability to generalize across directional flips and cutting ratios. 
As summarized in Figure~\ref{fig:transfer}, 
the model shows a significant degradation when trained on one configuration 
and evaluated on its directional or proportional counterpart.
When the model is trained on 0.25-ratio cuts from the right side and then evaluated on the mirrored left-side instruction (e.g., ‘cut the banana at a 0.25 ratio from the left’), its success rate drops from 55\% to 20\%.
Similarly, in the reverse direction (0.75 $\rightarrow$ 0.25), the success rate decreases from 35\% to 0\%, 
and when transferring from the 0.5-ratio to 0.25-ratio instruction, performance drops from 60\% to 25\%.
These findings demonstrate that even strong diffusion-based policies 
fail to generalize symmetrically across directional and ratio-based instructions. 
To bridge this gap, we introduce the ContinuousCut dataset, 
which provides dense ratio supervision (0.1–0.9) with explicit left/right labels. 
This enables continuous grounding of cutting geometry 
and fosters robust generalization across spatial and linguistic variations.
\paragraph{Numeric Ratio Grounding Failure in Continuous-Cut.}
As shown in Figure~\ref{fig:continuouscut}, our continuous-ratio benchmark exposes a systematic weakness of current VLA-style policies in grounding percentage-based ratios to spatial geometry. Even the large diffusion-based baseline (\textbf{RDT-1B}) behaves inconsistently across nearby targets: on \textit{banana}, success rate is $60\%$ at ratio $0.40$ cut but drops to $35\%$ at $0.70$ and $0.75$; on \textit{cucumber}, $60\%$ at ratio $0.25$ falls to $25\%$ at $0.75$. \textbf{OpenVLA} and \textbf{Octo} likewise fluctuate sharply and often collapse at mid-range ratios, as seen on banana at ratio $0.40$ where OpenVLA$5\%$ and Octo $0\%$. 
Thus, to correctly execute instructions such as “cut at 40\% from the right,”
the model must accurately link linguistic quantities to object-centric coordinates
and correctly perceive the object’s size and spatial proportions.

\subsection{Ablation Study}
\begin{figure}[ht]
    \centering
    \includegraphics[width=\linewidth]{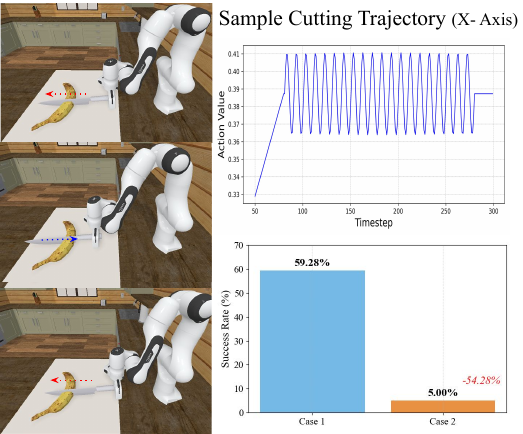}
    \caption{Visualization of task success rate for different trajectory styles.
\textbf{Case 1} represents the normal trajectory with the proposed Cutting Style Transfer Module (CSTM) applied, 
while \textbf{Case 2} corresponds to training on sawcut trajectories.}
    \label{fig:sawcut}
\vspace{-3mm}
\end{figure}

\paragraph{Effect of Fine-Grained Motion Styles and CSTM Module.}
As shown in Figure~\ref{fig:sawcut}, training on fine-grained, repetitive motions such as sawcut drastically reduces task success (59.28\% $\rightarrow$ 5.00\%) on RDT-1B model, as dense oscillatory trajectories distort motion feature distributions and destabilize policy learning. To address this, we propose the \textbf{Cutting Style Transfer Module (CSTM)}, which preserves the global trajectory structure while adaptively transferring local style cues. Rather than replicating high-frequency motions, CSTM regularizes trajectory dynamics to maintain temporal smoothness and consistency. Detailed qualitative results, including visualizations of predicted trajectory coordinates, are provided in the Supplementary Material.

\begin{figure}[ht]
    \centering
    \includegraphics[width=\linewidth]{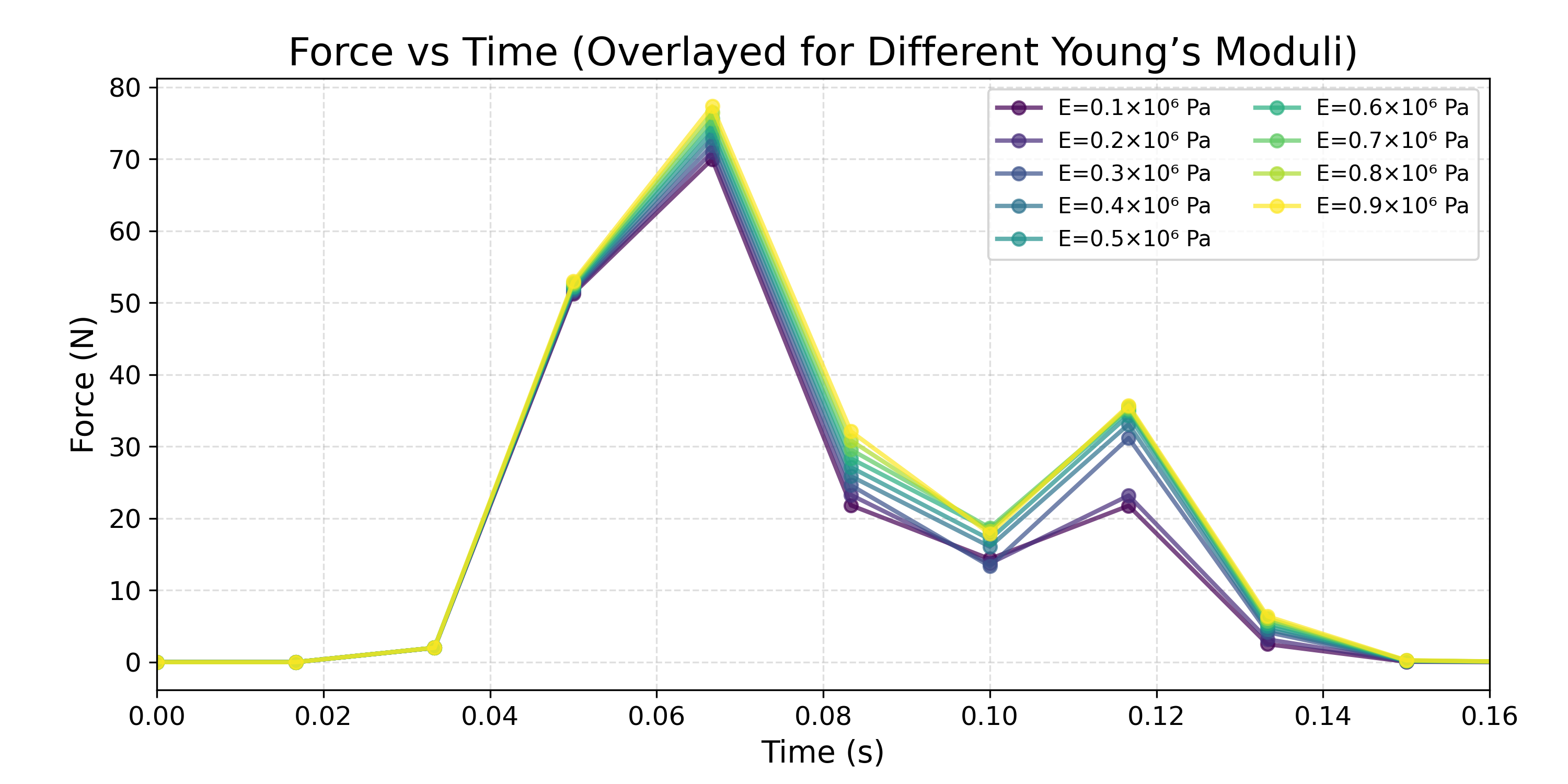}
    \caption{Temporal progression of knife force magnitude for all tested Young's moduli.}
    \label{fig:force_time_overlay}
\vspace{-3mm}
\end{figure}
\begin{figure}[ht!]
    \centering
    \includegraphics[width=\linewidth]{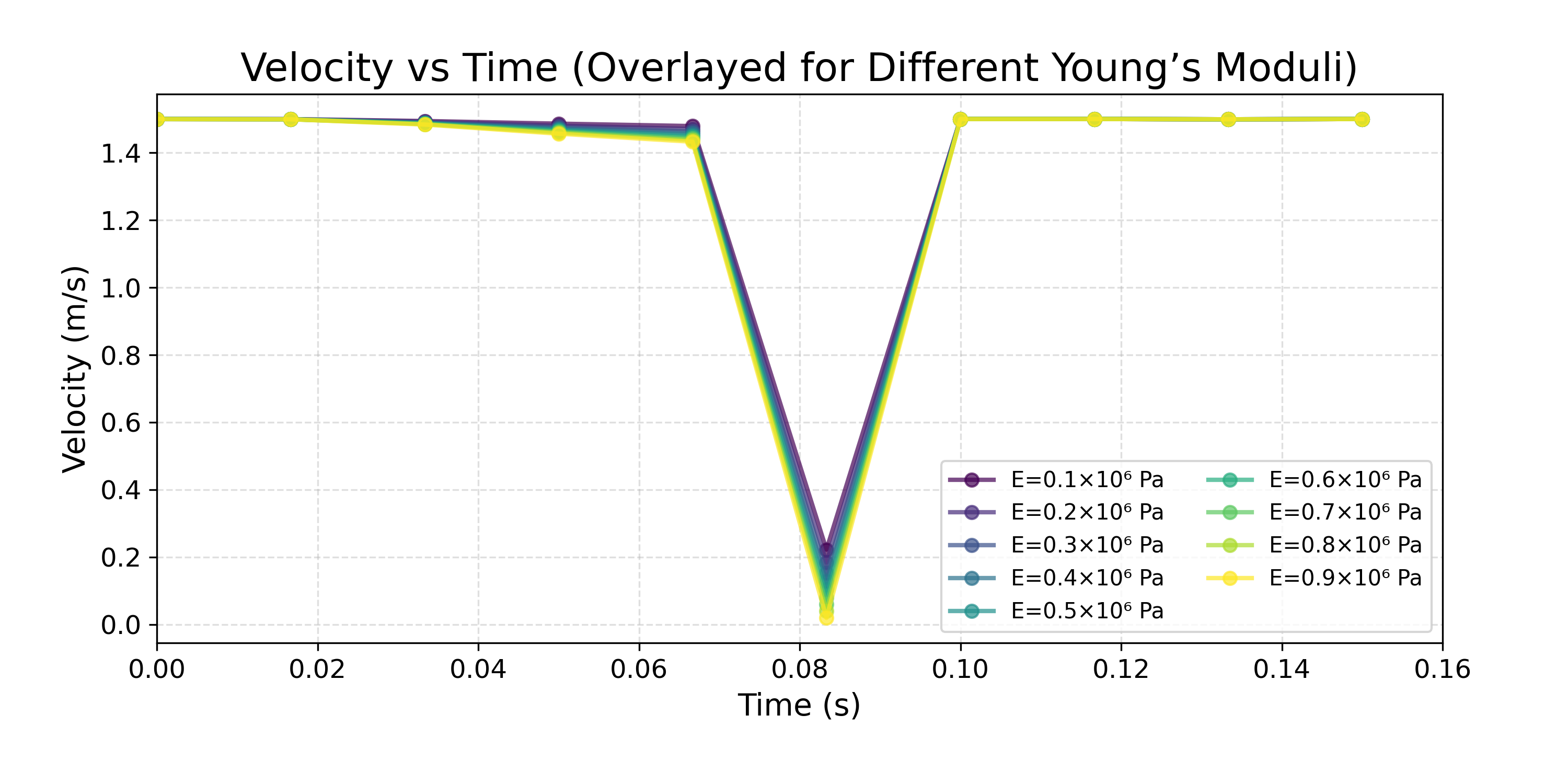}
    \caption{Temporal progression of knife velocity for all tested Young's moduli.}
    \label{fig:velocity_time_overlay}
\vspace{-3mm}
\end{figure}
\begin{figure}[ht]
    \centering
    \includegraphics[width=\linewidth]{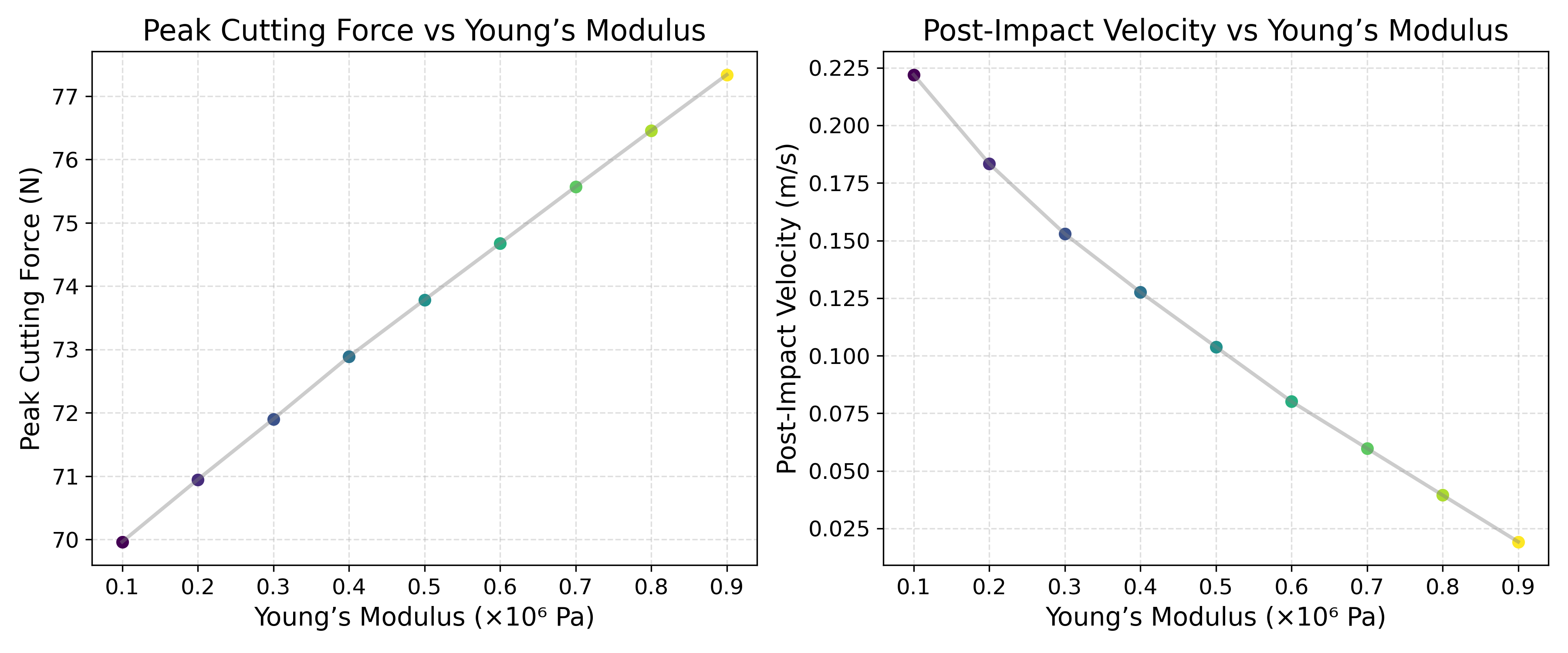}
    \caption{Comparison of peak cutting force and post-impact velocity across all tested Young's moduli.}
    \label{fig:youngs_moduli_summary}
\vspace{-3mm}
\end{figure}
\paragraph{Assessing the Physical Realism of MPM-Based Cutting Simulation}
To evaluate the physical realism of the MPM-based Cutting Simulation, the Young's Modulus \(E\) was systematically varied between \(0.1 \times 10^{6}~\mathrm{Pa}\) and \(0.9 \times 10^{6}~\mathrm{Pa}\), in increments of \(0.1 \times 10^{6}~\mathrm{Pa}\). This parameter sweep was designed to assess how material stiffness influences the peak cutting force and post-impact velocity during a single cutting cycle (downward stroke and retraction).

Figure~\ref{fig:force_time_overlay} shows the temporal progression of the knife force magnitude throughout the simulation for all nine tested Young's moduli. Each curve is color-coded from purple (lowest \(E\)) to yellow (highest \(E\)), illustrating that as stiffness increases, the knife experiences higher cutting forces. Figure~\ref{fig:velocity_time_overlay} similarly plots the knife velocity over time, using the same color scheme. The velocities reveal the opposite trend, demonstrating that stiffer materials lead to lower knife velocities.

The force magnitudes were directly obtained from the simulation output, while velocities were estimated via a finite-difference approximation of the time derivative of position, computed from the simulation output.

Figure~\ref{fig:youngs_moduli_summary} summarizes the results by comparing the recorded peak cutting force (\(\mathrm{N}\)) and post-impact velocity (\(\mathrm{m/s}\)) for each tested modulus. The peak force, defined as the maximum knife force across time for a given \(E\), consistently occurs at the fifth timestep for all simulations. Peak cutting force increases approximately linearly with \(E\), from \(69.96~\mathrm{N}\) at \(E = 0.1 \times 10^{6}~\mathrm{Pa}\) to \(77.34~\mathrm{N}\) at \(E = 0.9 \times 10^{6}~\mathrm{Pa}\). The post-impact velocity, defined as the minimum knife velocity following impact (occurring at the sixth timestep), shows a corresponding linear decrease, from \(0.22~\mathrm{m/s}\) at \(E = 0.1 \times 10^{6}~\mathrm{Pa}\) to \(0.02~\mathrm{m/s}\) at \(E = 0.9 \times 10^{6}~\mathrm{Pa}\). These results are physically consistent: stiffer materials (larger \(E\)) resist deformation more, requiring greater cutting forces and storing more elastic energy, which slows the blade after impact. The observed force-velocity relationship validates that our MPM simulation realistically captures the interplay between elastic stiffness, cutting resistance, and energy dissipation.
\section{Conclusion}
\label{sec:conclusion}
We propose a unified framework for vision-language-action-physics~(VLAP) in deformable object manipulation, centered on food cutting. 
By combining a physically realistic MLS-MPM simulator with a large-scale multimodal dataset, 
our approach enables quantitative evaluation of geometric and physical reasoning in cutting tasks. 
The \textbf{CulinaryCut} benchmark further enhances cutting performance through physically grounded learning.
\clearpage
\setcounter{page}{1}
\setcounter{section}{0}
\renewcommand{\thesection}{S\arabic{section}}
\appendix

\setcounter{figure}{0}
\setcounter{table}{0}
\renewcommand{\thefigure}{S\arabic{figure}}
\renewcommand{\thetable}{S\arabic{table}}
\section{Details on MLS-MPM Simulation Method for Cutting}
\label{sec:mpm}

Our simulation methodology is based on MPM.
MPM represents materials as collections of particles, or point clouds.
Each particle carries physical quantities such as mass, volume, position, and velocity.
On a given timestep of the simulation, the particle-based quantities are interpolated or rasterized to a background grid, often a uniform square/cubic Cartesian grid.
Then, governing equations such as conservation of momentum are solved on the grid, predicting updated values at the next time step.
These updated quantities are interpolated back to the particles, and the particles' positions are updated for the next time step.


In addition to each particle carrying position $\mathbf{x}$, velocity $\mathbf{v}$, and deformation gradient $\mathbf{F}$, in MLS-MPM, an affine velocity term $\mathbf{C}$ is also stored on each particle.
This increases the fidelity of velocities transferred between particles and the background grid---and hence, the fidelity of the overall simulation.
In our case, to simulate cutting, we also store on each particle an accumulated plasticity $\alpha$ and a continuum damage scalar $D\!\in\![0,1]$.
Our background grid is a uniform lattice with spacing $\Delta x$.

\subpara{Governing equations.}
At the continuum level, we enforce mass and linear momentum balance together with the kinematic relation for $\mathbf{F}$; the Cauchy stress $\boldsymbol{\sigma}$ is closed by a constitutive law:

\begin{subequations}
\label{eq:gov}
\begin{align}
\dot{\rho} + \rho\,\nabla\!\cdot\!\mathbf{v} &= 0, \label{eq:gov_mass}\\
\rho\,\dot{\mathbf{v}} &= \nabla\!\cdot\!\boldsymbol{\sigma} + \rho\,\mathbf{g}, \label{eq:gov_mom}\\
\dot{\mathbf{F}} &= (\nabla \mathbf{v})\,\mathbf{F}. \label{eq:gov_kin}
\end{align}
\end{subequations}

We adopt corotated elasticity as the elastic backbone and couple it with (optional) J2 plasticity via radial return. Around the blade we accumulate a scalar damage $D$ that linearly reduces the effective Lam\'e moduli, which promotes controlled crack opening.

\subpara{P2G: particle-to-grid transfer.}
Each particle distributes mass and momentum to its $3\!\times\!3\!\times\!3$ stencil using {quadratic} B-spline weights; the APIC/MLS affine term $\mathbf{C}$ is transferred to better preserve rotation/shear modes and reduce numerical diffusion. Stress is computed from the corotated model by polar decomposition $\mathbf{F}=\mathbf{R}\mathbf{S}$ (rotation $\mathbf{R}$; stretch $\mathbf{S}$). Damage acts by scaling the Lam\'e parameters, and when yielding occurs we apply J2 radial return (optionally with a small viscoplastic parameter) to tame overstress during abrupt contact transients. The Cauchy stress and the corotated first Piola stress are

\begin{subequations}
\label{eq:corot_simple}
\begin{align}
\boldsymbol{\sigma} &= J^{-1}\,\mathbf{P}\,\mathbf{F}^{\mathsf T},\\
\mathbf{P} &= 2\,\mu\,(\mathbf{F}-\mathbf{R}) + \lambda\,(J-1)\,J\,(\mathbf{F}^{-1})^{\mathsf T},\\
&\text{where } \mathbf{F}=\mathbf{R}\mathbf{S},\quad J=\det(\mathbf{F}).
\end{align}
\end{subequations}

\begin{equation}
\label{eq:lame}
\mu=\frac{E}{2(1+\nu)},\qquad
\lambda=\frac{E\,\nu}{(1+\nu)(1-2\nu)}.
\end{equation}

\subpara{Grid update.}
Grid node velocities are obtained by dividing momentum by mass, followed by gravity, boundary, and mild damping. Knife and board are represented as signed distance fields (SDFs). For nodes inside blade/board AABBs, we sample SDFs and resolve contact by decomposing relative velocity into normal/tangential components, applying non-penetration (normal restitution) and Coulomb friction; a CPIC-style conservative mix can be used to strengthen momentum consistency with the tool. We accumulate the impulse from pre- and post-contact velocities and convert it to an average force over the output-accumulation window $\Delta t_{\mathrm{acc}}$:
\begin{equation}
\label{eq:force}
\mathbf{F}_{\mathrm{avg}} \;=\;
\frac{\sum_i m_i\big(\mathbf{v}_i^{\mathrm{after}}-\mathbf{v}_i^{\mathrm{before}}\big)}{\Delta t_{\mathrm{acc}}}.
\end{equation}
This force directly underpins the force--time curves reported in Section \ref{sec:experiment}.
In parallel, we build a contact-strength scalar $\hat c\!\in\![0,1]$ by normalizing the accumulated approach speed near the blade with $c_{\mathrm{norm}}=0.35\,\Delta x/\Delta t$; $\hat c$ gates cutting (damage updates occur only when contact is sufficient) and drives a quadratic speed-resistance model for the knife: with normalized speed $u=s/s_0$, we update $u\leftarrow u/(1+k_2\,\hat c\,u\,\Delta t)$, scaling $k_2$ by material $(E,\sigma_y)$ so harder materials decelerate faster. Here, \(E\) denotes the Young’s modulus.

\subpara{G2P: grid-to-particle transfer.}
Updated grid velocities are interpolated back to particles to form the new particle velocity and affine term. We apply gravity, particle-level damping, and a small ``tip force'' to particles within a narrow band around the blade edge to gently separate the two sides and mitigate stick--slip oscillations. The deformation gradient is advanced with the affine field and clamped isotropically when $J$ leaves a safe interval:
\begin{equation}
\label{eq:Fupdate}
\begin{aligned}
\mathbf{F}^{n+1} &= \big(\mathbf{I} + \Delta t\,\mathbf{C}^{n+1}\big)\,\mathbf{F}^{n},\\
&\text{if } J\notin[J_{\min},J_{\max}] \text{ then } \mathbf{F}\leftarrow J^{-1/3}\mathbf{F}.
\end{aligned}
\end{equation}

\subpara{Damage-gated cutting and resolution-invariant scaling.}
Damage accumulation is enabled only within a blade SDF band $|\phi|<\text{band}$ when three conditions hold simultaneously: sufficient contact strength ($\hat c\ge\hat c_{\min}$), sufficiently fast approach ($v_n\le -v_{\mathrm{th}}$), and a downward stroke. As $D$ grows, effective stiffness decreases, and cracks open progressively; visual/topological separation is stabilized by connectivity-/color-based post-processing of particles. To preserve behavior across resolutions we define
{resolution-invariant} thresholds via 
$\text{band}=\text{band}_0(\Delta x_{\mathrm{ref}}/\Delta x)^\gamma$ and 
$v_{\mathrm{th}}=(\Delta x/\Delta t)\,\hat v$.

\subpara{Stabilization, efficiency, and reproducibility.}
We use a CFL-limited step ($\Delta t \le 0.2\,\Delta x/\sqrt{(\lambda+2\mu)/\rho}$), cap particle/grid speeds, clamp $J$, and apply AABB culling of blade/board SDF queries.
Implementation is on Taichi\citep{10.1145/3355089.3356506} GPU kernels with atomic impulse reductions and tuned block sizes.
For reproducibility we avoid logging full deformation fields and instead record a lightweight set of summaries: 
(i) force--time curves with a momentum-change check $\sum_i m_i\,\Delta\mathbf v_i \approx \int \mathbf F\,dt$, 
(ii) $J$ statistics (min/max and clamp counts), 
(iii) contact-strength $\hat c$ traces and blade/handle/board impulses, and (iv) configuration metadata ($\Delta x,\Delta t$, material parameters, knife/EEF states). 
When $\Delta x,\Delta t$ change we rescale \texttt{band}, $v_{\mathrm{th}}$, and $c_{\mathrm{norm}}$ as specified; when materials change we scale $k_2$ with $(E,\sigma_y)$.
\begin{figure*}[ht!]
    \centering
    \includegraphics[width=\linewidth]{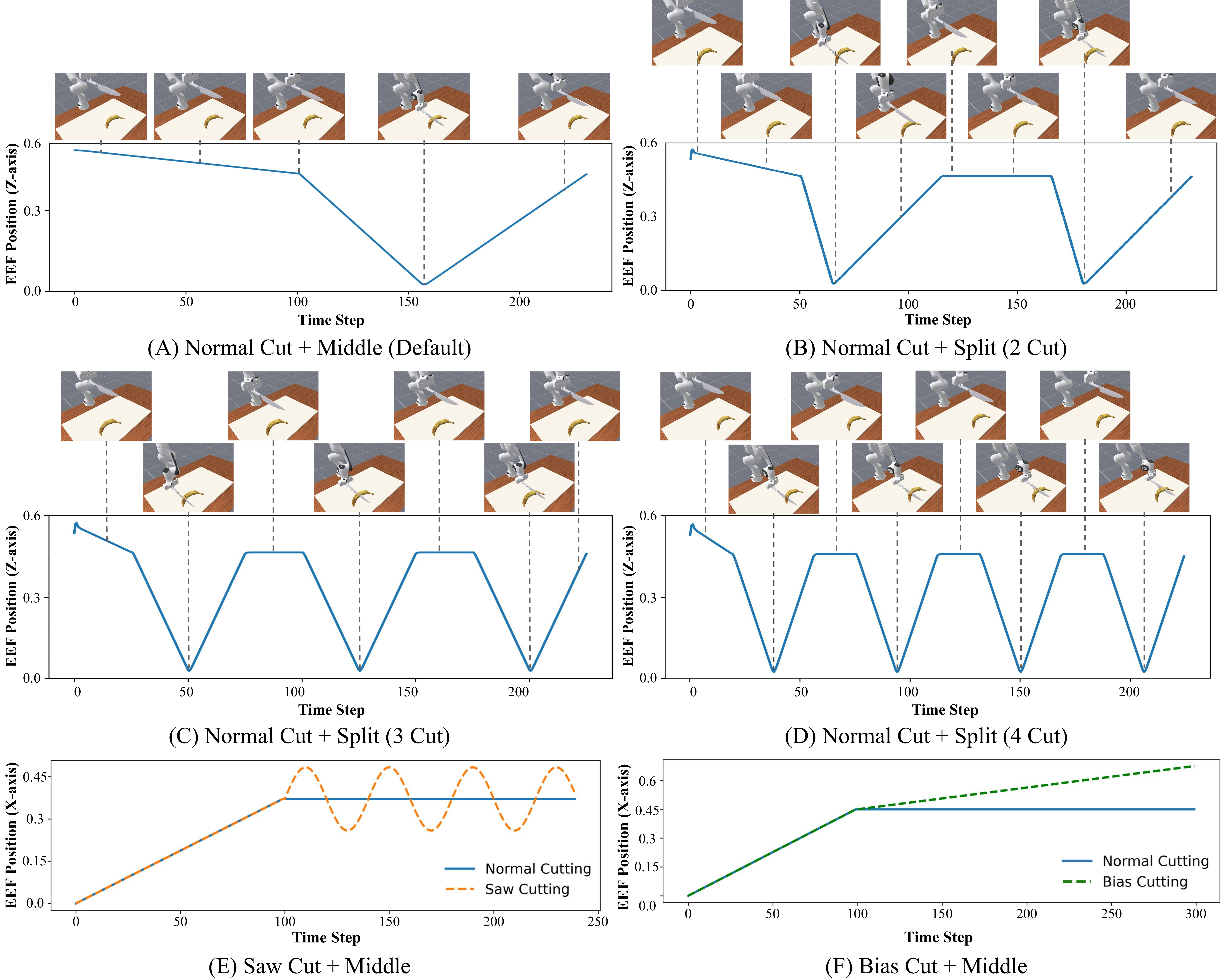}
    \caption{\textbf{Trajectory visualization.}~End-effector (EE) trajectory visualizations for different Cut Style + Continuous Cut State combination. (A) Normal Cut + Middle produces a single vertical descent toward the object’s midpoint. (B)--(D) Normal Cut + Split show multi-cut trajectories with 2-cut, 3-cut, and 4-cut sequences, each consisting of repeated vertical slicing motions at distinct target locations. (E) Saw Cut introduces periodic lateral oscillations characteristic of sawing actions. (F) Bias Cut exhibits a diagonally slanted descent produced by a continuous lateral shift during cutting.
}
    \label{fig:traj_vis}
\end{figure*}
\begin{table}[t!]
\centering
\resizebox{\linewidth}{!}{
\begin{tabular}{l|l|c}
\toprule
\textbf{Cut Style~$\ell_s$} & \textbf{Cut State} & \textbf{Representative Language Instruction} \\
\midrule
\hline

\textbf{Normal Cut} & Ratio Cut &
Normal Cut the~$O_t$ at 0.3 ratio from the right side. \\

\textbf{Normal Cut} & Middle Cut &
Normal Cut the~$O_t$ at the midpoint along its length. \\

\textbf{Normal Cut} & Split Cut &
Normal Cut the~$O_t$ at the first split boundary from the right side. \\
\midrule

\textbf{Bias Cut} & Ratio Cut &
Bias cut the~$O_t$ at 0.4 ratio toward the right end. \\

\textbf{Bias Cut} & Middle Cut &
Bias cut the~$O_t$ at the midpoint from the left side. \\

\textbf{Bias Cut} & Split Cut &
Bias cut the~$O_t$ at the first split boundary along its length. \\
\midrule

\textbf{Guillotine Cut} & Ratio Cut &
Guillotine cut the~$O_t$ at 0.5 ratio from the left side. \\

\textbf{Guillotine Cut} & Middle Cut &
Guillotine cut the~$O_t$ at the center from the top direction. \\

\textbf{Guillotine Cut} & Split Cut &
Guillotine cut the~$O_t$ at the first split boundary from the top. \\
\midrule

\textbf{Saw Cut} & Ratio Cut &
Saw cut the~$O_t$ at 0.6 ratio along its length. \\

\textbf{Saw Cut} & Middle Cut &
Saw cut the~$O_t$ at the midpoint from the right side. \\

\textbf{Saw Cut} & Split Cut &
Saw cut the~$O_t$ at the third boundary toward the right end. \\
\bottomrule
\end{tabular}
}
\caption{
Representative language instructions for all combinations of Cut Style (Normal, Bias, Guillotine, Saw)
and Continuous Cut State (Ratio, Middle, Split). 
}
\label{tab:lang_inst_samples}
\end{table}
\section{Language Instruction Design.}
\label{sec:lang_inst}

This section outlines the design of the language instructions used in the CulinaryCut dataset. 
Each instruction is paired with a single trajectory and corresponds to a specific Cutting Specification, defined as a combination of a \textit{Cut Style} and a \textit{Continuous Cut State}.

\paragraph{Default Settings.}
When an instruction specifies only a Continuous Cut State (Ratio, Middle, Split) and does not specify the Cut Style, we default to using the \textbf{Normal Cut} style.
Conversely, when only the Cut Style (Bias, Guillotine, Saw) is provided without a Continuous Cut State, we assign the \textbf{Middle Cut} state.
These default rules ensure that every instruction is complete and consistently resolves to a single, unambiguous Cutting Specification.

\paragraph{Design Objectives.}
Instructions are constructed to convey the information required for the Cutting Specification. 
For Continuous Cut States, we include numerical terms, geometric references, or count-based expressions as needed. 
We augment each template with additional paraphrases generated by large language models to increase linguistic diversity while preserving the intended semantics.

\paragraph{Instruction Structure.}
Each instruction consists of four components:
\begin{itemize}[leftmargin=*]
    \item \textbf{Target Object~$O_t$:} banana, cucumber, apple, peach, melon, orange, strawberry
    \item \textbf{Cut Style~$\ell_s$:} Normal, Bias, Guillotine, Saw
    \item \textbf{Continuous Cut State:} Ratio, Middle, Split
    \item \textbf{Directional Reference:} from the left/right side, from the top, along its length
\end{itemize}

\paragraph{Instruction Samples.}
\autoref{tab:lang_inst_samples} provides representative instruction examples for all combinations of Cut Style and Continuous Cut State. 
Each entry illustrates how an instruction encodes both the cutting motion and the target position. 
When an instruction omits either component, we apply the default rules described above.

\section{Trajectory Visualization.}
\label{sec:supp_traj_vis}

To highlight the structural differences among cutting strategies, \autoref{fig:traj_vis} presents representative end-effector (EE) trajectories extracted from the dataset. 
All trajectories are plotted in the world coordinate frame and temporally aligned, enabling direct comparison across different Cut Style~$\times$~Continuous Cut State combinations.

\paragraph{Normal Cut.}
Normal Cutting exhibits a straight downward motion pattern, characterized by a monotonic decrease along the Z-axis with minimal lateral movement. 
In the Middle Cut case (\autoref{fig:traj_vis} (A)), the EE performs a single descent toward the geometric center of the object.
For Split Cuts (\autoref{fig:traj_vis} (B--D)), the trajectory contains multiple vertical segments, each corresponding to a distinct slicing operation in the multi-cut sequence. 
Despite the increased number of cuts, the vertical descent profile remains consistent, reflecting the stable and non-oscillatory nature of Normal Cutting.
In the Split Cut settings, increasing the number of cuts simply adds additional vertical descent segments, each executed at a new target location. 
In practice, the 2-cut, 3-cut, and 4-cut trajectories differ only in the number of repeated descending motions, while the underlying slicing dynamics remain identical across all cases.

\paragraph{Saw Cut.}
Saw Cutting introduces periodic lateral oscillations superimposed on the downward motion (\autoref{fig:traj_vis} (E)).
These oscillations appear as sinusoidal variations along the X-axis and represent the forward–backward micro-motions characteristic of sawing.
The oscillation frequency is randomly sampled at the start of each demonstration, resulting in diverse but consistently periodic patterns while maintaining the overall sawing behavior.

\paragraph{Bias Cut.}
Bias Cutting exhibits a diagonal slicing motion produced by a continuous lateral drift during descent.
As shown in \autoref{fig:traj_vis}, the EE trajectory deviates steadily from the straight-line profile of the Normal Cut, forming an oblique motion path.
This lateral shift results in an angled contact geometry, aligning with the definition of Bias Cutting in ~Sec.~\ref{Subsec:task_define}.

\section{Additional Experiment Result}
\subsection{Additional Result}
\begin{table}[h]
\centering
\caption{\textbf{Distribution of successfully placed cuts in Split Cut evaluation with RDT-1B.}~Each cell shows the percentage of episodes in which exactly $k$ cuts were placed within the $\pm 10\%$ tolerance among all required cuts for that Split Cut configuration.}

\label{tab:splitcut}
\begin{tabular}{c|c|cccc}
\toprule
Object & Split Cut & 1 & 2 & 3 & 4 \\\midrule
\multirow{3}{*}{Banana}
  & 1/3 & 20\% & 0\%  & --  & --  \\
  & 1/4 & 10\% & 5\%  & 0\% & --  \\
  & 1/5 & 5\%  & 0\%  & 0\% & 0\% \\
\midrule
\multirow{3}{*}{Cucumber}
  & 1/3 & 30\% & 0\%  & --  & --  \\
  & 1/4 & 10\% & 0\%  & 0\% & --  \\
  & 1/5 & 0\%  & 0\%  & 0\% & 0\% \\
\bottomrule
\end{tabular}
\end{table}

\paragraph{Split Cut Evaluation.}
To analyze how VLA policies behave in multi-segment cutting, we consider the Split Cut setting, where the agent must place multiple cuts so that an object is divided into equal-length pieces.
We treat an episode as \emph{successful} only if all required cuts fall within a $\pm 10\%$ tolerance around their ideal boundaries.
Under this strict criterion, the overall Split Cut success rate is \textbf{0\%} for all baselines (RDT-1B, Octo, and OpenVLA), indicating that none of the models can complete a full cutting sequence reliably.

\autoref{tab:splitcut} provides a finer-grained view of partial success for RDT-1B.
For each object and Split Cut configuration (1/3, 1/4, 1/5), the columns labeled 1–4 report the percentage of episodes in which exactly that many cuts were placed correctly among all required cuts.
For example, in the 1/3 Split Cut task (which requires two cuts), the entry \textbf{20\%} in the “1” column for banana indicates that 20\% of episodes achieved exactly one correct cut and never both; the “2” column remains at 0\%, showing that no episode successfully placed both boundaries.
In the more challenging 1/4 setting (three required cuts), banana still reaches a small number of partially successful episodes: 10\% of episodes achieve exactly one correct cut, and 5\% manage to place two correct cuts, but no episode reaches all three.
Cucumber generally shows slightly higher rates of obtaining at least one correct cut (e.g., 30\% for the 1/3 setting), yet still fails to complete the full sequence under any configuration.

These distributions reveal a characteristic failure mode of long-horizon cutting.
Once the first cut is misplaced, the policy rarely reorganizes the remaining cuts to recover the global pattern.
Instead, the controller tends to hover around a local region and repeatedly issue similar cutting motions, producing strange, oscillatory trajectories rather than progressing to new boundaries.
In other words, the models can sometimes “hit” one boundary, or occasionally two in the 1/4 setting, but they do not learn stable multi-step cutting programs.
This helps explain why partial statistics in \autoref{tab:splitcut} are non-zero, while the strict all-cuts-complete success remains 0\% for RDT-1B, Octo, and OpenVLA alike.

\paragraph{Ablation Study of Topology Update.}
\begin{table}[t!]
    \centering
    \caption{Average Cut Completion by Object with RDT\textendash1B Models}
    \resizebox{\columnwidth}{!}{
    \begin{tabular}{c|c|c|c}
    \toprule
    Case & Train Topology  & Evaluation Topology & Success Rate \\
    \midrule
    Case 1 &  update \xmark & update \xmark  & 55\% \\
    Case 2 &   update \xmark & update \cmark & 50\% \\
    Case 3 & update \cmark& update \xmark & 45\% \\
    Case 4(ours) & update \cmark& update \cmark& 60\% \\
    \bottomrule
    \end{tabular}
    }
    \label{fig:ablation_topology}
\end{table}
\begin{figure}[t!]
    \centering
    \includegraphics[width=\linewidth]{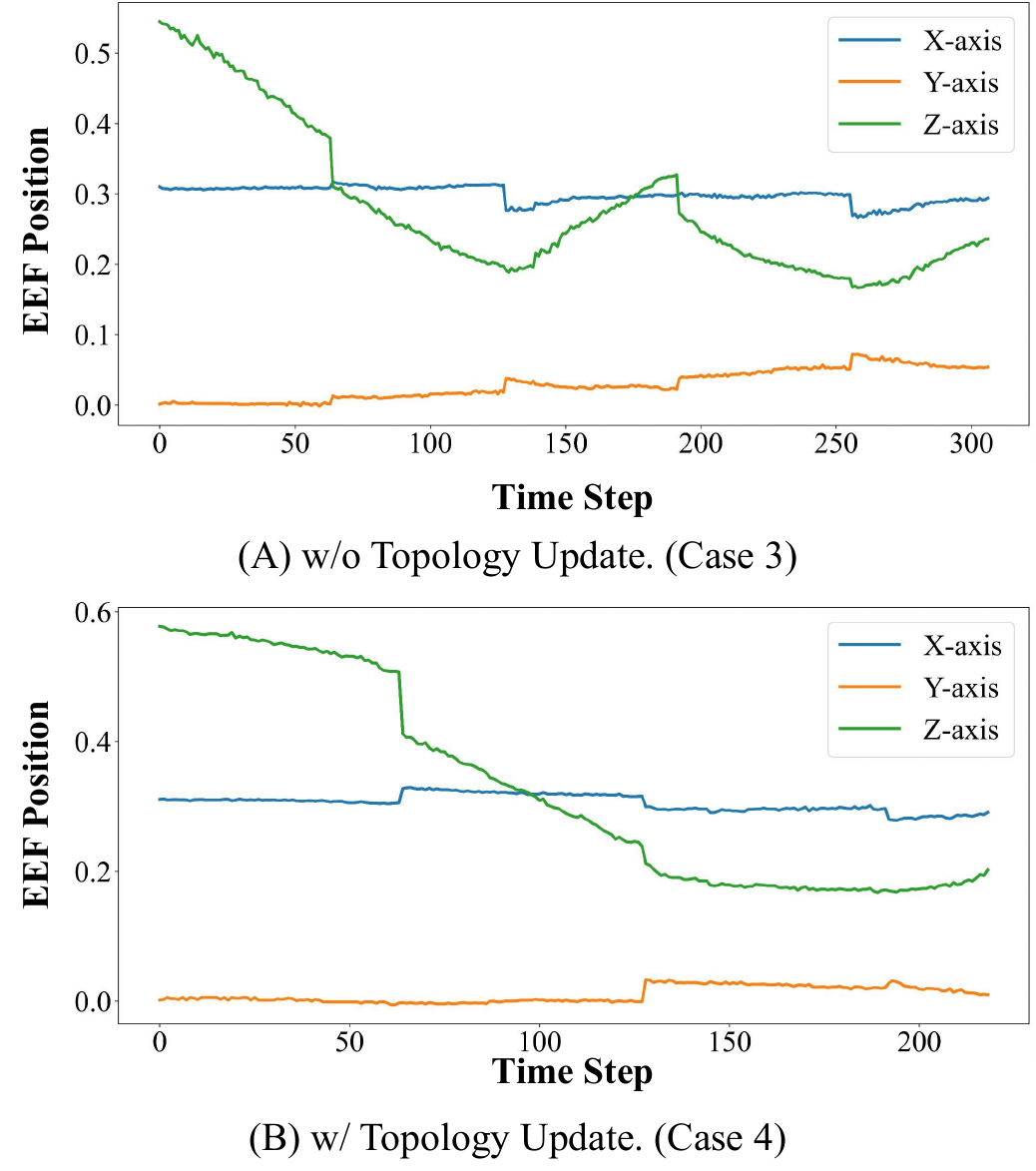}
    \caption{\textbf{Predicted 3D end-effector trajectories from RDT under different topology-update conditions.}~(A) Without topology update (Case 3), the predicted trajectories show larger axis-wise deviations. (B) With topology update (Case 4), the trajectories become more stable as cutting progresses, demonstrating the importance of modeling topology changes.}
    \label{fig:wotopoupdate}
\end{figure}
\autoref{fig:ablation_topology} reports an ablation on Train Topology and Evaluation Topology, where we independently control whether the object mesh is allowed to update its connectivity after a cut. 
Here, ``update \cmark'' denotes that the mesh connectivity is updated once the cut is completed, so that the rendered observations explicitly show separated parts, visible gaps, and other topology changes. 
In contrast, ``update \xmark'' corresponds to a topology-free setting in which the mesh connectivity is kept fixed even after the knife passes through the object, and only the surface geometry is deformed.
The four rows in \autoref{fig:ablation_topology} correspond to all combinations of training and evaluation conditions.
Quantitatively, the average success rates vary only moderately across these settings: the matched regimes in Case~1 and Case~4 achieve 55\% and 60\%, while the mismatched regimes in Case~2 and Case~3 drop to 50\% and 45\%, respectively. 
However, these relatively small differences in average cut completion mask systematic qualitative changes in the behavior of the RDT\textendash 1B.

When the model is trained on topology-aware data but evaluated in an environment where the topology is not updated (Case~3), it cannot reliably perceive that a cut has already succeeded. 
As a result, the policy frequently returns to nearly the same 3D pose and executes a second, redundant cutting motion along an almost identical trajectory, as visualized in \autoref{fig:wotopoupdate} (A). 

In contrast, when the same model is deployed in an environment that updates the object topology after each successful cut (Case~4), the change in geometry is clearly reflected in the visual input, and the policy typically performs a single decisive cut and then terminates or moves on to the next instruction, as shown in \autoref{fig:wotopoupdate} (B). 
A similar form of distribution shift also appears in Case~2, where the model is trained without topology updates but evaluated with them, leading to slightly degraded average performance compared to the fully topology-aware configuration in Case~4.

Although these experiments are conducted in simulation, where we can artificially switch topology updates on or off, real-world cutting inevitably changes the object topology and this change is immediately visible to the robot's camera. 
Therefore, even though the numerical differences in \autoref{fig:ablation_topology} are modest, we regard topology-aware data generation and evaluation (Case~4) as an important design choice for obtaining stable, non-redundant cutting behavior in real robotic systems.
\begin{figure}[t!]
    \centering
    \includegraphics[width=\linewidth]{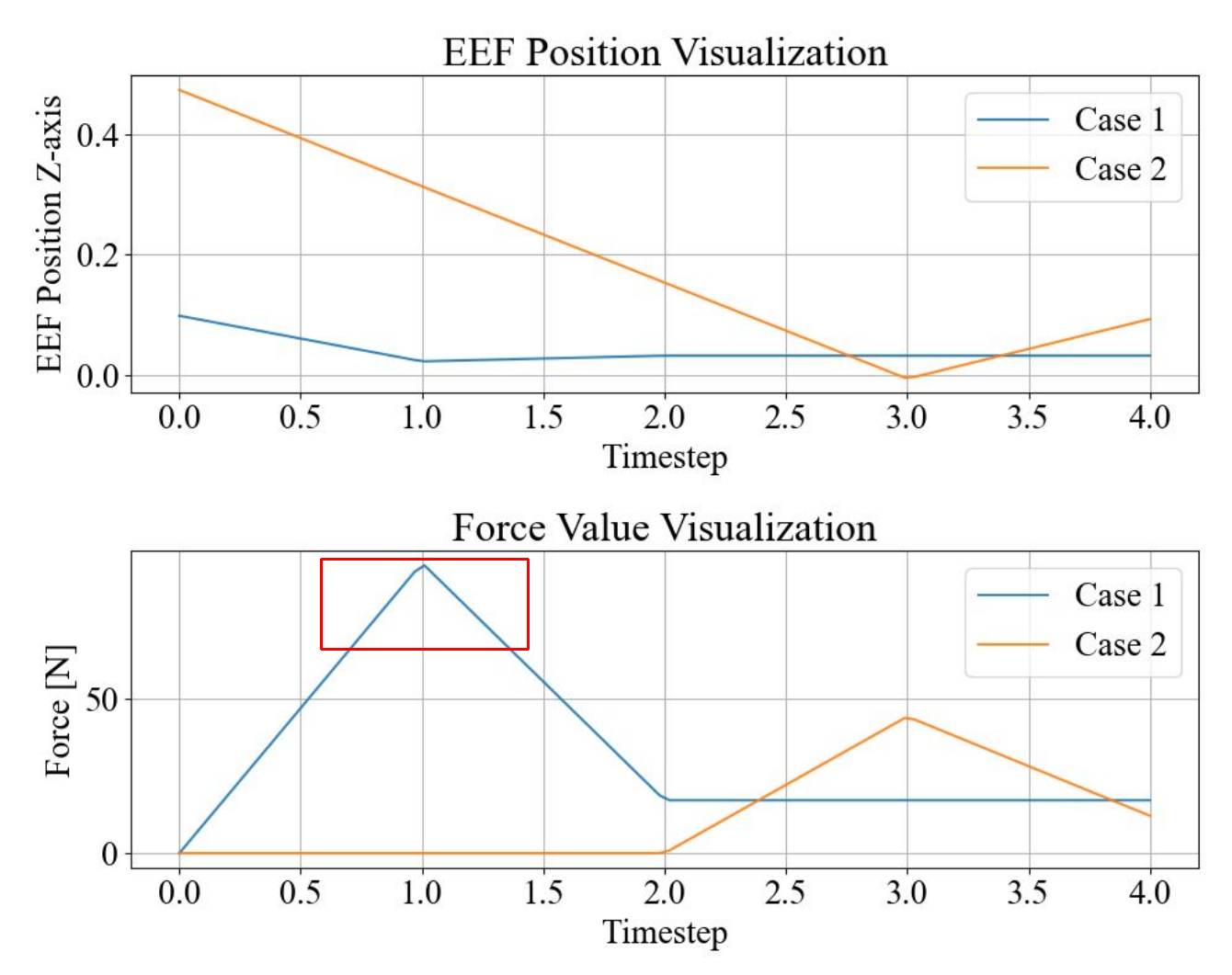}
    \caption{\textbf{Visualization of the end-effector (EEF) trajectory and 
    contact force during the cutting motion.} The top plot illustrates the EEF 
    position along the cutting direction, while the bottom plot shows the 
    corresponding contact force profiles. \textbf{Case~1} denotes the model 
    \emph{without} the Safety Module, and \textbf{Case~2} represents the model 
    \emph{with} the proposed Safety Module enabled. For clarity, regions where 
    Case~1 exhibits excessive force (exceeding 60\,N) are highlighted in 
    \textcolor{red}{red}, indicating unsafe spikes that the Safety Module 
    successfully suppresses.
    }
    \label{fig:mpmsimdata}
\end{figure}
\paragraph{Ablation Study of Safety module.}
\begin{table}[t!]
    \centering
    \caption{Average maximum cutting speed before and after applying the proposed safety module.}
    \resizebox{\columnwidth}{!}{
    \begin{tabular}{c|c|c|c}
    \toprule
    Safety Module & Maximum cutting speed  & Maximum force  \\
    \midrule
    Safety Module \xmark &  3.63 m/s  & 129.31N  \\
    Safety Module \cmark &   0.89 m/s  & 37.78N \\
    \bottomrule
    \end{tabular}
    }
    \label{Tab:Safety}
\end{table}
To validate the effectiveness of our manipulation safety module, we compare the maximum cutting velocity and contact force before and after applying the proposed limiter. As shown in \autoref{Tab:Safety} and \autoref{fig:mpmsimdata}, the baseline model without the safety module exhibits an excessively aggressive motion pattern: across 20 trials, the predicted average maximum cutting velocity reaches 3.63\,m/s, and the corresponding peak contact force in the physics simulation increases to 129.31\,N, surpassing the Franka robot’s safe operational limit of 100\,N.

In contrast, when the proposed safety module is enabled, the commanded velocity is adaptively regulated according to contact conditions. As visualized in \autoref{Tab:Safety}, the resulting execution reduces the average maximum cutting speed to 0.89\,m/s, while the simulated peak contact force remains safely contained at 37.78\,N.

These results indicate that the proposed safety module effectively suppresses force spikes that may otherwise cause hardware damage, producing material-aware and stable cutting trajectories. Even under challenging contact interactions, the module ensures physically safe manipulation without compromising task feasibility.

\clearpage
{
    \small
    \bibliographystyle{ieeenat_fullname}
    \bibliography{main}

@String(CVPR= {IEEE Conf. Comput. Vis. Pattern Recog.})

@String(ICCV= {Int. Conf. Comput. Vis.})

@String(ECCV= {Eur. Conf. Comput. Vis.})

@String(CVPR  = {CVPR})

@String(ICCV  = {ICCV})

@String(ECCV  = {ECCV})

@article{achiam2023gpt4,
  title   = {GPT-4 Technical Report},
  author  = {OpenAI},
  journal = {arXiv preprint arXiv:2303.08774},
  year    = {2023}
}

@inproceedings{radford2021clip,
  title     = {Learning Transferable Visual Models From Natural Language Supervision},
  author    = {Radford, Alec and Kim, Jong Wook and Hallacy, Chris and Ramesh, Aditya and Goh, Gabriel and Agarwal, Sandhini and Sastry, Girish and Askell, Amanda and Mishkin, Pamela and Clark, Jack and Krueger, Gretchen and Sutskever, Ilya},
  booktitle = {Proceedings of the 38th International Conference on Machine Learning (ICML)},
  series    = {PMLR},
  volume    = {139},
  pages     = {8748--8763},
  year      = {2021}
}

@article{liu2023llava,
  title   = {Visual Instruction Tuning},
  author  = {Liu, Haotian and Li, Chunyuan and Li, Yuheng and Lee, Yong Jae},
  journal = {arXiv preprint arXiv:2304.08485},
  year    = {2023}
}

@article{zhu2023minigpt4,
  title   = {MiniGPT-4: Enhancing Vision-Language Understanding with Advanced Large Language Models},
  author  = {Zhu, Deyao and Chen, Jun and Shen, Xiaoqian and Li, Xiang and Elhoseiny, Mohamed},
  journal = {arXiv preprint arXiv:2304.10592},
  year    = {2023}
}

@article{cai2023vipllava,
  title   = {ViP-LLaVA: Making Large Multimodal Models Understand Arbitrary Visual Prompts},
  author  = {Cai, Mu and Liu, Haotian and Park, Dennis and Mustikovela, Siva Karthik and Meyer, Gregory P. and Chai, Yuning and Lee, Yong Jae},
  journal = {arXiv preprint arXiv:2312.00784},
  year    = {2023}
}

@inproceedings{rasheed2024glamm,
  title     = {GLaMM: Pixel Grounding Large Multimodal Model},
  author    = {Rasheed, Hanoona and Maaz, Muhammad and Mullappilly, Sahal Shaji and Shaker, Abdelrahman and Khan, Salman and Cholakkal, Hisham and Anwer, Rao M. and Xing, Eric and Yang, Ming-Hsuan and Khan, Fahad Shahbaz},
  booktitle = {Proceedings of the IEEE/CVF Conference on Computer Vision and Pattern Recognition (CVPR)},
  year      = {2024}
}

@article{li2024llavainterleave,
  title   = {LLaVA-NeXT-Interleave: Tackling Multi-image, Video, and 3D in Large Multimodal Models},
  author  = {Li, Feng and Zhang, Renrui and Zhang, Hao and Zhang, Yuanhan and Li, Bo and Li, Wei and Ma, Zejun and Li, Chunyuan},
  journal = {arXiv preprint arXiv:2407.07895},
  year    = {2024}
}

@inproceedings{liu2024mmbench,
  title     = {MMBench: Is Your Multi-modal Model an All-around Player?},
  author    = {Liu, Yulin and Li, Bo and Li, Chunyuan and others},
  booktitle = {Proceedings of the European Conference on Computer Vision (ECCV)},
  year      = {2024}
}

@article{li2023seedbench2,
  title   = {SEED-Bench-2: Benchmarking Multimodal Large Language Models},
  author  = {Li, Bohao and Ge, Yuying and Ge, Yixiao and Wang, Guangzhi and Wang, Rui and Zhang, Ruimao and Shan, Ying},
  journal = {arXiv preprint arXiv:2311.17092},
  year    = {2023}
}

@article{fang2024mmbenchvideo,
  title   = {MMBench-Video: A Long-Form Multi-Shot Benchmark for Holistic Video Understanding},
  author  = {Fang, Xinyu and Mao, Kangrui and Duan, Haodong and Zhao, Xiangyu and Li, Yining and Lin, Dahua and Chen, Kai},
  journal = {arXiv preprint arXiv:2406.14515},
  year    = {2024}
}

@article{openxembodiment2023,
  title   = {Open X-Embodiment: Robotic Learning Datasets and RT-X Models},
  author  = {Open X-Embodiment Collaboration and O'Neill, Abby and Rehman, Abdul and Gupta, Abhinav and et al.},
  journal = {arXiv preprint arXiv:2310.08864},
  year    = {2023},
  url     = {https://arxiv.org/abs/2310.08864},
  doi     = {10.48550/arXiv.2310.08864}
}

@inproceedings{robonet2019,
  title     = {RoboNet: Large-Scale Multi-Robot Learning},
  author    = {Dasari, Sudeep and Ebert, Frederik and Tian, Stephen and Nair, Suraj and Bucher, Ben and Schmeckpeper, Karl and Singh, Siddharth and Levine, Sergey and Finn, Chelsea},
  booktitle = {Proceedings of the Conference on Robot Learning (CoRL)},
  year      = {2019},
  url       = {https://arxiv.org/abs/1910.11215}
}

@article{bridgedata2023,
  title   = {BridgeData V2: A Dataset for Robot Learning at Scale},
  author  = {Walke, Homer and Black, Kevin and Lee, Abraham and Kim, Moo Jin and Du, Max and Zheng, Chongyi and Zhao, Tony and Hansen-Estruch, Philippe and Vuong, Quan and He, Andre and Myers, Vivek and Fang, Kuan and Finn, Chelsea and Levine, Sergey},
  journal = {arXiv preprint arXiv:2308.12952},
  year    = {2023},
  url     = {https://arxiv.org/abs/2308.12952}
}

@article{droid2024,
  title   = {DROID: A Large-Scale In-The-Wild Robot Manipulation Dataset},
  author  = {Khazatsky, Alexander and et al.},
  journal = {arXiv preprint arXiv:2403.12945},
  year    = {2024},
  url     = {https://arxiv.org/abs/2403.12945}
}

@inproceedings{libero2023,
  title     = {LIBERO: Benchmarking Knowledge Transfer for Lifelong Robot Learning},
  author    = {Liu, Bang and Wang, Zhe and Ding, Zhi and others},
  booktitle = {Advances in Neural Information Processing Systems (NeurIPS) — Datasets and Benchmarks Track},
  year      = {2023},
  url       = {https://proceedings.neurips.cc/paper_files/paper/2023/file/8c3c666820ea055a77726d66fc7d447f-Paper-Datasets_and_Benchmarks.pdf}
}

@article{disect2022,
  title   = {DiSECt: A Differentiable Simulator for Parameter Inference and Control in Robotic Cutting},
  author  = {Heiden, Eric and Macklin, Miles and Narang, Yashraj and Fox, Dieter and Garg, Animesh and Ramos, Fabio},
  journal = {arXiv preprint arXiv:2203.10263},
  year    = {2022},
  url     = {https://arxiv.org/abs/2203.10263}
}

@article{sliceit2024,
  title   = {A Dual Simulator Framework for Learning Robot Food Slicing},
  author  = {Beltran-Hernandez, Carlos C. and others},
  journal = {arXiv preprint arXiv:2404.02569},
  year    = {2024},
  url     = {https://arxiv.org/abs/2404.02569}
}

@article{topocut2025,
  title   = {TopoCut: Learning Multi-Step Cutting with Spectral Rewards and Discrete Diffusion Policies},
  author  = {Wang, Liquan and Bian, Jiangjie and Heiden, Eric and Garg, Animesh},
  journal = {arXiv preprint arXiv:2509.19712},
  year    = {2025},
  url     = {https://arxiv.org/abs/2509.19712}
}

@article{openvla,
  title   = {OpenVLA: An Open-Source Vision-Language-Action Model},
  author  = {Kim, Moo Jin and Pertsch, Karl and Karamcheti, Siddharth and Xiao, Ted and Balakrishna, Ashwin and Nair, Suraj and Rafailov, Rafael and Foster, Ethan and Lam, Grace and Sanketi, Pannag and Vuong, Quan and Kollar, Thomas and Burchfiel, Benjamin and Tedrake, Russ and Sadigh, Dorsa and Levine, Sergey and Liang, Percy and Finn, Chelsea},
  journal = {arXiv preprint arXiv:2406.09246},
  year    = {2024},
  url     = {https://arxiv.org/abs/2406.09246}
}

@article{sulsky1994particle,
    doi = {10.1016/0045-7825(94)90112-0},
    url = {https://www.sciencedirect.com/science/article/pii/0045782594901120},
    title = {{A Particle Method for History-Dependent Materials}},
    journal = {{Computer Methods in Applied Mechanics and Engineering}},
    volume = {118},
    number = {1},
    pages = {179-196},
    year = {1994},
    issn = {0045-7825},
    author = {D. Sulsky and Z. Chen and H.L. Schreyer}
}

@article{jiang2017angular,
    url = {https://www.sciencedirect.com/science/article/pii/S0021999117301535},
    doi = {10.1016/j.jcp.2017.02.050},
    title = {{A}n {A}ngular {M}omentum {C}onserving {A}ffine-{P}article-in-{C}ell {M}ethod},
    author = {Chenfanfu Jiang and Craig Schroeder and Joseph Teran},
    journal = {Journal of {C}omputational {P}hysics},
    volume = {338},
    pages = {137-164},
    year = {2017},
    issn = {0021-9991}
}

@article{2018-MLSMPM,
    url = {https://doi.org/10.1145/3197517.3201293},
    doi = {10.1145/3197517.3201293},
    author = {Hu, Yuanming and Fang, Yu and Ge, Ziheng and Qu, Ziyin and Zhu, Yixin and Pradhana, Andre and Jiang, Chenfanfu},
    title = {{A} {M}oving {L}east {S}quares {M}aterial {P}oint {M}ethod with {D}isplacement {D}iscontinuity and {T}wo-Way {R}igid {B}ody {C}oupling},
    year = {2018},
    issue_date = {August 2018},
    publisher = {{A}ssociation for {C}omputing {M}achinery},
    address = {New York, NY, USA},
    volume = {37},
    number = {4},
    issn = {0730-0301},
    journal = {ACM Trans. Graph.},
    month = {jul},
    articleno = {150},
    numpages = {14}
}

@article{2013-mpm,
    url = {https://doi.org/10.1145/2461912.2461948},
    doi = {10.1145/2461912.2461948},
    author = {Stomakhin, Alexey and Schroeder, Craig and Chai, Lawrence and Teran, Joseph and Selle, Andrew},
    title = {{A} {M}aterial {P}oint {M}ethod for {S}now {S}imulation},
    year = {2013},
    issue_date = {July 2013},
    publisher = {Association for Computing Machinery},
    address = {New York, NY, USA},
    volume = {32},
    number = {4},
    issn = {0730-0301},
    journal = {ACM Trans. Graph.},
    month = {7},
    articleno = {102},
    numpages = {10}
}

@article{2019-cdmpm,
    url = {https://doi.org/10.1145/3306346.3322949},
    doi = {10.1145/3306346.3322949},
    author = {Wolper, Joshuah and Fang, Yu and Li, Minchen and Lu, Jiecong and Gao, Ming and Jiang, Chenfanfu},
    title = {{CD-MPM}: {C}ontinuum {D}amage {M}aterial {P}oint {M}ethods for {D}ynamic {F}racture {A}nimation},
    year = {2019},
    issue_date = {August 2019},
    publisher = {Association for Computing Machinery},
    address = {New York, NY, USA},
    volume = {38},
    number = {4},
    issn = {0730-0301},
    journal = {ACM Trans. Graph.},
    month = {jul},
    articleno = {119},
    numpages = {15}
}

@article{10.1145/3340259,
    author = {Wang, Stephanie and Ding, Mengyuan and Gast, Theodore F. and Zhu, Leyi and Gagniere, Steven and Jiang, Chenfanfu and Teran, Joseph M.},
    title = {Simulation and Visualization of Ductile Fracture with the Material Point Method},
    year = {2019},
    issue_date = {July 2019},
    publisher = {Association for Computing Machinery},
    address = {New York, NY, USA},
    volume = {2},
    number = {2},
    url = {https://doi.org/10.1145/3340259},
    doi = {10.1145/3340259},
    journal = {Proc. ACM Comput. Graph. Interact. Tech.},
    month = jul,
    articleno = {18},
    numpages = {20}
}

@inproceedings{jamdagni2021robotic,
  title={Robotic slicing of fruits and vegetables: modeling the effects of fracture toughness and knife geometry},
  author={Jamdagni, Prajjwal and Jia, Yan-Bin},
  booktitle={2021 IEEE International Conference on Robotics and Automation (ICRA)},
  pages={6607--6613},
  year={2021},
  organization={IEEE}
}

@article{dikshit2023robochop,
  title={Robochop: Autonomous framework for fruit and vegetable chopping leveraging foundational models},
  author={Dikshit, Atharva and Bartsch, Alison and George, Abraham and Farimani, Amir Barati},
  journal={arXiv preprint arXiv:2307.13159},
  year={2023}
}

@article{shi2023robocook,
  title={Robocook: Long-horizon elasto-plastic object manipulation with diverse tools},
  author={Shi, Haochen and Xu, Huazhe and Clarke, Samuel and Li, Yunzhu and Wu, Jiajun},
  journal={arXiv preprint arXiv:2306.14447},
  year={2023}
}

@article{luo2025fmb,
  title={Fmb: a functional manipulation benchmark for generalizable robotic learning},
  author={Luo, Jianlan and Xu, Charles and Liu, Fangchen and Tan, Liam and Lin, Zipeng and Wu, Jeffrey and Abbeel, Pieter and Levine, Sergey},
  journal={The International Journal of Robotics Research},
  volume={44},
  number={4},
  pages={592--606},
  year={2025},
  publisher={SAGE Publications Sage UK: London, England}
}

@inproceedings{sawhney2020playing,
  title={Playing with food: Learning food item representations through interactive exploration},
  author={Sawhney, Amrita and Lee, Steven and Zhang, Kevin and Veloso, Manuela and Kroemer, Oliver},
  booktitle={International Symposium on Experimental Robotics},
  pages={309--322},
  year={2020},
  organization={Springer}
}

@article{wang2023mimicplay,
  title={Mimicplay: Long-horizon imitation learning by watching human play},
  author={Wang, Chen and Fan, Linxi and Sun, Jiankai and Zhang, Ruohan and Fei-Fei, Li and Xu, Danfei and Zhu, Yuke and Anandkumar, Anima},
  journal={arXiv preprint arXiv:2302.12422},
  year={2023}
}

@article{shah2023mutex,
  title={Mutex: Learning unified policies from multimodal task specifications},
  author={Shah, Rutav and Mart{\'\i}n-Mart{\'\i}n, Roberto and Zhu, Yuke},
  journal={arXiv preprint arXiv:2309.14320},
  year={2023}
}

@article{heo2025furniturebench,
  title={Furniturebench: Reproducible real-world benchmark for long-horizon complex manipulation},
  author={Heo, Minho and Lee, Youngwoon and Lee, Doohyun and Lim, Joseph J},
  journal={The International Journal of Robotics Research},
  volume={44},
  number={10-11},
  pages={1863--1891},
  year={2025},
  publisher={SAGE Publications Sage UK: London, England}
}

@inproceedings{gong2023arnold,
  title={ARNOLD: A Benchmark for Language-Grounded Task Learning With Continuous States in Realistic 3D Scenes},
  author={Gong, Ran and Huang, Jiangyong and Zhao, Yizhou and Geng, Haoran and Gao, Xiaofeng and Wu, Qingyang and Ai, Wensi and Zhou, Ziheng and Terzopoulos, Demetri and Zhu, Song-Chun and others},
  booktitle={Proceedings of the IEEE/CVF International Conference on Computer Vision (ICCV)},
  year={2023}
}

@misc{wu2025robomindbenchmarkmultiembodimentintelligence,
      title={RoboMIND: Benchmark on Multi-embodiment Intelligence Normative Data for Robot Manipulation}, 
      author={Kun Wu and Chengkai Hou and Jiaming Liu and Zhengping Che and Xiaozhu Ju and Zhuqin Yang and Meng Li and Yinuo Zhao and Zhiyuan Xu and Guang Yang and Shichao Fan and Xinhua Wang and Fei Liao and Zhen Zhao and Guangyu Li and Zhao Jin and Lecheng Wang and Jilei Mao and Ning Liu and Pei Ren and Qiang Zhang and Yaoxu Lyu and Mengzhen Liu and Jingyang He and Yulin Luo and Zeyu Gao and Chenxuan Li and Chenyang Gu and Yankai Fu and Di Wu and Xingyu Wang and Sixiang Chen and Zhenyu Wang and Pengju An and Siyuan Qian and Shanghang Zhang and Jian Tang},
      year={2025},
      eprint={2412.13877},
      archivePrefix={arXiv},
      primaryClass={cs.RO},
      url={https://arxiv.org/abs/2412.13877}, 
}

@inproceedings{
wu2024unleashing,
title={Unleashing Large-Scale Video Generative Pre-training for Visual Robot Manipulation},
author={Hongtao Wu and Ya Jing and Chilam Cheang and Guangzeng Chen and Jiafeng Xu and Xinghang Li and Minghuan Liu and Hang Li and Tao Kong},
booktitle={The Twelfth International Conference on Learning Representations},
year={2024},
url={https://openreview.net/forum?id=NxoFmGgWC9}
}

@inproceedings{
li2024visionlanguage,
title={Vision-Language Foundation Models as Effective Robot Imitators},
author={Xinghang Li and Minghuan Liu and Hanbo Zhang and Cunjun Yu and Jie Xu and Hongtao Wu and Chilam Cheang and Ya Jing and Weinan Zhang and Huaping Liu and Hang Li and Tao Kong},
booktitle={The Twelfth International Conference on Learning Representations},
year={2024},
url={https://openreview.net/forum?id=lFYj0oibGR}
}

@misc{kim2024openvlaopensourcevisionlanguageactionmodel,
      title={OpenVLA: An Open-Source Vision-Language-Action Model}, 
      author={Moo Jin Kim and Karl Pertsch and Siddharth Karamcheti and Ted Xiao and Ashwin Balakrishna and Suraj Nair and Rafael Rafailov and Ethan Foster and Grace Lam and Pannag Sanketi and Quan Vuong and Thomas Kollar and Benjamin Burchfiel and Russ Tedrake and Dorsa Sadigh and Sergey Levine and Percy Liang and Chelsea Finn},
      year={2024},
      eprint={2406.09246},
      archivePrefix={arXiv},
      primaryClass={cs.RO},
      url={https://arxiv.org/abs/2406.09246}, 
}

@misc{zhou2025opendrivevlaendtoendautonomousdriving,
      title={OpenDriveVLA: Towards End-to-end Autonomous Driving with Large Vision Language Action Model}, 
      author={Xingcheng Zhou and Xuyuan Han and Feng Yang and Yunpu Ma and Alois C. Knoll},
      year={2025},
      eprint={2503.23463},
      archivePrefix={arXiv},
      primaryClass={cs.CV},
      url={https://arxiv.org/abs/2503.23463}, 
}

@misc{geminiroboticsteam2025geminiroboticsbringingai,
      title={Gemini Robotics: Bringing AI into the Physical World}, 
      author={Gemini Robotics Team and Saminda Abeyruwan and Joshua Ainslie and Jean-Baptiste Alayrac and Montserrat Gonzalez Arenas and Travis Armstrong and others},
      year={2025},
      eprint={2503.20020},
      archivePrefix={arXiv},
      primaryClass={cs.RO},
      url={https://arxiv.org/abs/2503.20020}, 
}

@misc{khan2025shakevlavisionlanguageactionmodelbasedbimanual,
      title={Shake-VLA: Vision-Language-Action Model-Based System for Bimanual Robotic Manipulations and Liquid Mixing}, 
      author={Muhamamd Haris Khan and Selamawit Asfaw and Dmitrii Iarchuk and Miguel Altamirano Cabrera and Luis Moreno and Issatay Tokmurziyev and Dzmitry Tsetserukou},
      year={2025},
      eprint={2501.06919},
      archivePrefix={arXiv},
      primaryClass={cs.RO},
      url={https://arxiv.org/abs/2501.06919}, 
}

@misc{li2025pointvlainjecting3dworld,
      title={PointVLA: Injecting the 3D World into Vision-Language-Action Models}, 
      author={Chengmeng Li and Junjie Wen and Yan Peng and Yaxin Peng and Feifei Feng and Yichen Zhu},
      year={2025},
      eprint={2503.07511},
      archivePrefix={arXiv},
      primaryClass={cs.RO},
      url={https://arxiv.org/abs/2503.07511}, 
}

@misc{yang2025agenticrobotbraininspiredframework,
      title={Agentic Robot: A Brain-Inspired Framework for Vision-Language-Action Models in Embodied Agents}, 
      author={Zhejian Yang and Yongchao Chen and Xueyang Zhou and Jiangyue Yan and Dingjie Song and Yinuo Liu and Yuting Li and Yu Zhang and Pan Zhou and Hechang Chen and Lichao Sun},
      year={2025},
      eprint={2505.23450},
      archivePrefix={arXiv},
      primaryClass={cs.RO},
      url={https://arxiv.org/abs/2505.23450}, 
}

@article{bu2025univla,
  title={UniVLA: Learning to Act Anywhere with Task-centric Latent Actions}, 
  author={Qingwen Bu and Yanting Yang and Jisong Cai and Shenyuan Gao and Guanghui Ren and Maoqing Yao and Ping Luo and Hongyang Li},
  journal={arXiv preprint arXiv:2505.06111},
  year={2025}
}

@misc{xiang2025vlamodelexpertcollaborationbidirectional,
      title={VLA Model-Expert Collaboration for Bi-directional Manipulation Learning}, 
      author={Tian-Yu Xiang and Ao-Qun Jin and Xiao-Hu Zhou and Mei-Jiang Gui and Xiao-Liang Xie and Shi-Qi Liu and Shuang-Yi Wang and Sheng-Bin Duang and Si-Cheng Wang and Zheng Lei and Zeng-Guang Hou},
      year={2025},
      eprint={2503.04163},
      archivePrefix={arXiv},
      primaryClass={cs.RO},
      url={https://arxiv.org/abs/2503.04163}, 
}

@misc{reed2022generalistagent,
      title={A Generalist Agent}, 
      author={Scott Reed and Konrad Zolna and Emilio Parisotto and Sergio Gomez Colmenarejo and Alexander Novikov and Gabriel Barth-Maron and Mai Gimenez and Yury Sulsky and Jackie Kay and Jost Tobias Springenberg and Tom Eccles and Jake Bruce and Ali Razavi and Ashley Edwards and Nicolas Heess and Yutian Chen and Raia Hadsell and Oriol Vinyals and Mahyar Bordbar and Nando de Freitas},
      year={2022},
      eprint={2205.06175},
      archivePrefix={arXiv},
      primaryClass={cs.AI},
      url={https://arxiv.org/abs/2205.06175}, 
}

@misc{pertsch2025fastefficientactiontokenization,
      title={FAST: Efficient Action Tokenization for Vision-Language-Action Models}, 
      author={Karl Pertsch and Kyle Stachowicz and Brian Ichter and Danny Driess and Suraj Nair and Quan Vuong and Oier Mees and Chelsea Finn and Sergey Levine},
      year={2025},
      eprint={2501.09747},
      archivePrefix={arXiv},
      primaryClass={cs.RO},
      url={https://arxiv.org/abs/2501.09747}, 
}

@misc{jiang2023vimageneralrobotmanipulation,
      title={VIMA: General Robot Manipulation with Multimodal Prompts}, 
      author={Yunfan Jiang and Agrim Gupta and Zichen Zhang and Guanzhi Wang and Yongqiang Dou and Yanjun Chen and Li Fei-Fei and Anima Anandkumar and Yuke Zhu and Linxi Fan},
      year={2023},
      eprint={2210.03094},
      archivePrefix={arXiv},
      primaryClass={cs.RO},
      url={https://arxiv.org/abs/2210.03094}, 
}

@misc{bucker2022lattelanguagetrajectorytransformer,
      title={LATTE: LAnguage Trajectory TransformEr}, 
      author={Arthur Bucker and Luis Figueredo and Sami Haddadin and Ashish Kapoor and Shuang Ma and Sai Vemprala and Rogerio Bonatti},
      year={2022},
      eprint={2208.02918},
      archivePrefix={arXiv},
      primaryClass={cs.RO},
      url={https://arxiv.org/abs/2208.02918}, 
}

@misc{brohan2023rt2visionlanguageactionmodelstransfer,
      title={RT-2: Vision-Language-Action Models Transfer Web Knowledge to Robotic Control}, 
      author={Anthony Brohan},
      year={2023},
      eprint={2307.15818},
      archivePrefix={arXiv},
      primaryClass={cs.RO},
      url={https://arxiv.org/abs/2307.15818}, 
}

@misc{lynch2022interactivelanguagetalkingrobots,
      title={Interactive Language: Talking to Robots in Real Time}, 
      author={Corey Lynch and Ayzaan Wahid and Jonathan Tompson and Tianli Ding and James Betker and Robert Baruch and Travis Armstrong and Pete Florence},
      year={2022},
      eprint={2210.06407},
      archivePrefix={arXiv},
      primaryClass={cs.RO},
      url={https://arxiv.org/abs/2210.06407}, 
}

@misc{huang2022innermonologueembodiedreasoning,
      title={Inner Monologue: Embodied Reasoning through Planning with Language Models}, 
      author={Wenlong Huang and Fei Xia and Ted Xiao and Harris Chan and Jacky Liang and Pete Florence and Andy Zeng and Jonathan Tompson and Igor Mordatch and Yevgen Chebotar and Pierre Sermanet and Noah Brown and Tomas Jackson and Linda Luu and Sergey Levine and Karol Hausman and Brian Ichter},
      year={2022},
      eprint={2207.05608},
      archivePrefix={arXiv},
      primaryClass={cs.RO},
      url={https://arxiv.org/abs/2207.05608}, 
}

@misc{huang2023instruct2actmappingmultimodalityinstructions,
      title={Instruct2Act: Mapping Multi-modality Instructions to Robotic Actions with Large Language Model}, 
      author={Siyuan Huang and Zhengkai Jiang and Hao Dong and Yu Qiao and Peng Gao and Hongsheng Li},
      year={2023},
      eprint={2305.11176},
      archivePrefix={arXiv},
      primaryClass={cs.RO},
      url={https://arxiv.org/abs/2305.11176}, 
}

@misc{gbagbe2024bivlavisionlanguageactionmodelbasedbimanual,
      title={Bi-VLA: Vision-Language-Action Model-Based System for Bimanual Robotic Dexterous Manipulations}, 
      author={Koffivi Fidèle Gbagbe and Miguel Altamirano Cabrera and Ali Alabbas and Oussama Alyunes and Artem Lykov and Dzmitry Tsetserukou},
      year={2024},
      eprint={2405.06039},
      archivePrefix={arXiv},
      primaryClass={cs.RO},
      url={https://arxiv.org/abs/2405.06039}, 
}

@misc{octomodelteam2024octoopensourcegeneralistrobot,
      title={Octo: An Open-Source Generalist Robot Policy}, 
      author={Octo Model Team and Dibya Ghosh and Homer Walke and Karl Pertsch and Kevin Black and Oier Mees and Sudeep Dasari and Joey Hejna and Tobias Kreiman and Charles Xu and Jianlan Luo and You Liang Tan and Lawrence Yunliang Chen and Pannag Sanketi and Quan Vuong and Ted Xiao and Dorsa Sadigh and Chelsea Finn and Sergey Levine},
      year={2024},
      eprint={2405.12213},
      archivePrefix={arXiv},
      primaryClass={cs.RO},
      url={https://arxiv.org/abs/2405.12213}, 
}

@misc{zhang2025upvlaunifiedunderstandingprediction,
      title={UP-VLA: A Unified Understanding and Prediction Model for Embodied Agent}, 
      author={Jianke Zhang and Yanjiang Guo and Yucheng Hu and Xiaoyu Chen and Xiang Zhu and Jianyu Chen},
      year={2025},
      eprint={2501.18867},
      archivePrefix={arXiv},
      primaryClass={cs.CV},
      url={https://arxiv.org/abs/2501.18867}, 
}

@misc{ajay2023compositionalfoundationmodelshierarchical,
      title={Compositional Foundation Models for Hierarchical Planning}, 
      author={Anurag Ajay and Seungwook Han and Yilun Du and Shuang Li and Abhi Gupta and Tommi Jaakkola and Josh Tenenbaum and Leslie Kaelbling and Akash Srivastava and Pulkit Agrawal},
      year={2023},
      eprint={2309.08587},
      archivePrefix={arXiv},
      primaryClass={cs.LG},
      url={https://arxiv.org/abs/2309.08587}, 
}

@misc{liu2025rdt1bdiffusionfoundationmodel,
      title={RDT-1B: a Diffusion Foundation Model for Bimanual Manipulation}, 
      author={Songming Liu and Lingxuan Wu and Bangguo Li and Hengkai Tan and Huayu Chen and Zhengyi Wang and Ke Xu and Hang Su and Jun Zhu},
      year={2025},
      eprint={2410.07864},
      archivePrefix={arXiv},
      primaryClass={cs.RO},
      url={https://arxiv.org/abs/2410.07864}, 
}

@article{zhang2025inspire,
  title={InSpire: Vision-Language-Action Models with Intrinsic Spatial Reasoning},
  author={Zhang, Ji and Wu, Shihan and Luo, Xu and Wu, Hao and Gao, Lianli and Shen, Heng Tao and Song, Jingkuan},
  journal={arXiv preprint arXiv:2505.13888},
  year={2025}
}

@misc{wu2025foresightforethoughtvlminthelooppolicy,
      title={From Foresight to Forethought: VLM-In-the-Loop Policy Steering via Latent Alignment}, 
      author={Yilin Wu and Ran Tian and Gokul Swamy and Andrea Bajcsy},
      year={2025},
      eprint={2502.01828},
      archivePrefix={arXiv},
      primaryClass={cs.RO},
      url={https://arxiv.org/abs/2502.01828}, 
}

@article{dreamvla25,
          author = {Wenyao Zhang and
                    Hongsi Liu and
                    Zekun Qi and
                    Yunan Wang and
                    Xinqiang Yu and
                    Jiazhao Zhang and
                    Runpei Dong and
                    Jiawei He and
                    He Wang and
                    Zhizheng Zhang and
                    Li Yi and 
                    Wenjun Zeng and
                    Xin Jin},
          title        = {DreamVLA: A Vision-Language-Action Model Dreamed with Comprehensive World Knowledge},
          journal      = {CoRR},
          volume       = {abs/2507.04447},
          year         = {2025},
          url          = {https://doi.org/10.48550/arXiv.2507.04447},
          doi          = {10.48550/ARXIV.2507.04447},
          eprinttype    = {arXiv},
          eprint       = {2507.04447}
        }

@article{yu2025forcevla,
  title={{ForceVLA}: Enhancing {VLA} Models with a Force-aware {MoE} for Contact-rich Manipulation},
  author={Yu, Jiawen and Liu, Hairuo and Yu, Qiaojun and Ren, Jieji and Hao, Ce and Ding, Haitong and Huang, Guangyu and Huang, Guofan and Song, Yan and Cai, Panpan and others},
  journal={arXiv preprint arXiv:2505.22159},
  year={2025}
}

@inproceedings{xia2018gibson,
  title={Gibson env: Real-world perception for embodied agents},
  author={Xia, Fei and Zamir, Amir R and He, Zhiyang and Sax, Alexander and Malik, Jitendra and Savarese, Silvio},
  booktitle={Proceedings of the IEEE conference on computer vision and pattern recognition},
  pages={9068--9079},
  year={2018}
}

@misc{makoviychuk2021isaacgymhighperformance,
      title={Isaac Gym: High Performance GPU-Based Physics Simulation For Robot Learning}, 
      author={Viktor Makoviychuk and Lukasz Wawrzyniak and Yunrong Guo and Michelle Lu and Kier Storey and Miles Macklin and David Hoeller and Nikita Rudin and Arthur Allshire and Ankur Handa and Gavriel State},
      year={2021},
      eprint={2108.10470},
      archivePrefix={arXiv},
      primaryClass={cs.RO},
      url={https://arxiv.org/abs/2108.10470}, 
}

@misc{tao2025maniskill3gpuparallelizedrobotics,
      title={ManiSkill3: GPU Parallelized Robotics Simulation and Rendering for Generalizable Embodied AI}, 
      author={Stone Tao and Fanbo Xiang and Arth Shukla and Yuzhe Qin and Xander Hinrichsen and Xiaodi Yuan and Chen Bao and Xinsong Lin and Yulin Liu and Tse-kai Chan and Yuan Gao and Xuanlin Li and Tongzhou Mu and Nan Xiao and Arnav Gurha and Viswesh Nagaswamy Rajesh and Yong Woo Choi and Yen-Ru Chen and Zhiao Huang and Roberto Calandra and Rui Chen and Shan Luo and Hao Su},
      year={2025},
      eprint={2410.00425},
      archivePrefix={arXiv},
      primaryClass={cs.RO},
      url={https://arxiv.org/abs/2410.00425}, 
}

@misc{huang2025tactilevlaunlockingvisionlanguageactionmodels,
      title={Tactile-VLA: Unlocking Vision-Language-Action Model's Physical Knowledge for Tactile Generalization}, 
      author={Jialei Huang and Shuo Wang and Fanqi Lin and Yihang Hu and Chuan Wen and Yang Gao},
      year={2025},
      eprint={2507.09160},
      archivePrefix={arXiv},
      primaryClass={cs.RO},
      url={https://arxiv.org/abs/2507.09160}, 
}

@article{yang2023learning,
            title={Learning Interactive Real-World Simulators},
            author={Yang, Mengjiao and Du, Yilun and Ghasemipour, Kamyar and Tompson, Jonathan and Schuurmans, Dale and Abbeel, Pieter},
            journal={arXiv preprint arXiv:2310.06114},
            year={2023}
            }

@misc{niu2024screenagentvisionlanguagemodeldriven,
      title={ScreenAgent: A Vision Language Model-driven Computer Control Agent}, 
      author={Runliang Niu and Jindong Li and Shiqi Wang and Yali Fu and Xiyu Hu and Xueyuan Leng and He Kong and Yi Chang and Qi Wang},
      year={2024},
      eprint={2402.07945},
      archivePrefix={arXiv},
      primaryClass={cs.HC},
      url={https://arxiv.org/abs/2402.07945}, 
}

@article{Xu_2025,
   title={A Differentiable Material Point Method Framework for Shape Morphing},
   volume={31},
   ISSN={2160-9306},
   url={http://dx.doi.org/10.1109/TVCG.2025.3591729},
   DOI={10.1109/tvcg.2025.3591729},
   number={10},
   journal={IEEE Transactions on Visualization and Computer Graphics},
   publisher={Institute of Electrical and Electronics Engineers (IEEE)},
   author={Xu, Michael and Song, Chang-Yong and Levin, David and Hyde, David},
   year={2025},
   month=oct, pages={9140–9153} }

@misc{khazatsky2024droidlargescaleinthewildrobot,
      title={DROID: A Large-Scale In-The-Wild Robot Manipulation Dataset}, 
      author={Alexander Khazatsky et al},
      year={2024},
      eprint={2403.12945},
      archivePrefix={arXiv},
      primaryClass={cs.RO},
      url={https://arxiv.org/abs/2403.12945}, 
}

@article{10.1145/3355089.3356506,
author = {Hu, Yuanming and Li, Tzu-Mao and Anderson, Luke and Ragan-Kelley, Jonathan and Durand, Fr\'{e}do},
title = {Taichi: a language for high-performance computation on spatially sparse data structures},
year = {2019},
issue_date = {December 2019},
publisher = {Association for Computing Machinery},
address = {New York, NY, USA},
volume = {38},
number = {6},
issn = {0730-0301},
url = {https://doi.org/10.1145/3355089.3356506},
doi = {10.1145/3355089.3356506},
journal = {ACM Trans. Graph.},
month = nov,
articleno = {201},
numpages = {16}
}
}
\end{document}